\documentclass{article}
    \PassOptionsToPackage{numbers,sort&compress}{natbib}

\usepackage[preprint]{neurips_2020}

\usepackage[utf8]{inputenc} 
\usepackage[T1]{fontenc}    
\usepackage[colorlinks,linkcolor=black,citecolor=black,urlcolor=black]{hyperref}       
\usepackage{url}            
\usepackage{booktabs}       
\usepackage{amsfonts}       
\usepackage{nicefrac}       
\usepackage{microtype}      

\usepackage{graphicx, caption}
\usepackage{color}
\usepackage{subfig}
\usepackage{etoolbox}

\input{./Definitions}
\graphicspath{{./figs/}}

\renewcommand{\citet}{\cite}
\renewcommand{\citep}{\cite}

\title{Proximal Mapping for Deep Regularization}

\author{%
  Mao Li, \qquad Yingyi Ma, \qquad Xinhua Zhang \\
  Department of Computer Science, 
  University of Illinois at Chicago\\
  Chicago, IL 60607 \\
  \texttt{\{mli206, yma36, zhangx\}@uic.edu} \\
}

\begin{document}

\maketitle

\vspace{-0.5em}
\begin{abstract}
  Underpinning the success of deep learning is effective regularizations that allow a variety of priors in data to be modeled.
  For example,
  robustness to adversarial perturbations, 
  and correlations between multiple modalities.  
  However, most regularizers are specified in terms of hidden layer outputs,
  which are not themselves optimization variables.
  In contrast to prevalent methods that optimize them indirectly through model weights,
  we propose inserting proximal mapping as a new layer to the deep network,
  which directly and explicitly produces well regularized hidden layer outputs.
  The resulting technique is shown well connected to kernel warping and dropout,
  and novel algorithms were developed for robust temporal learning and multiview modeling, 
  both outperforming state-of-the-art methods.
\end{abstract}

\vspace{-0.8em}
\section{Introduction}
\label{sec:introduction}
\vspace{-0.3em}


The success of deep learning relies on massive neural networks that often considerably out-scale the training dataset, defying the conventional learning theory \citep{ZhaBenHaretal17,Arpitzetal17}.
Regularization has been shown essential and a variety of forms are available.
For example, 
invariances to transformations such as rotation \citep{SimLecDen12} have been extended beyond group-based diffeomorphisms
to indecipherable transformations that are only exemplified by pairs of views \citep{PalKanAraSav17}, \eg, sentences uttered by the same person. 
Prior regularities are also commonly available
\textbf{a)} \emph{within} layers of neural networks,
such as sparsity \citep{MakFre14}, spatial invariance in convolutional nets, 
structured gradient that accounts for data covariance \citep{RotLucNowHof18}; 
%
\textbf{b)} \emph{between} layers of representation, 
such as stability under dropout and adversarial perturbations of preceding layers \citep{SriHinKrietal14}, 
contractivity between layers \citep{Rifaietal11}, 
and correlations in hidden layers among multiple views \citep{WanAroLivetal15,AndAroBiletal13};
and \textbf{c)} at batch level, 
\eg, disentangled representation and multiple modalities.

The most prevalent approach to incorporating priors is regularization,
which leads to the standard regularized risk minimization (RRM) for a given dataset $\Dcal$,
empirical distribution $\ptil$, and loss $\ell$:
\begin{align}
\label{eq:obj_RRM}
    \min\nolimits_f \ \EE_{x \sim \ptil} [\ell(f(x))] + \Gamma(f) + \sum\nolimits_i \Omega_i(\{h_i(x, f)\}_{x \in \Dcal}).
\end{align}
Here $f$ is the predictor (\eg, neural network),
and $\Gamma$ is the data-independent regularizer (\eg, $L_2$ norm), 
and $\Omega_i$ is the data-dependent regularizer on the $i$-th layer output $h_i$ under $f$
(\eg, invariance of $h_i$ with respect to the $i$-th step input $x_i$ in an RNN).
Note $\Omega_i$ can involve multiple layers (\eg, contractivity), or be decomposed over training examples.
Optimization techniques such as end-to-end training have produced strong performance,
along with progresses in the global analysis of the solution \citep[\eg,][]{Duetal19}.
However, all these analyses make assumptions on the landscape of the objective function,
which, although often satisfied by the empirical risk $\EE_{x \sim \ptil} [\ell(f(x))]$,
are typically violated or complicated by the addition of data-dependant regularizers $\Omega_i$.
The nontrivial contention between accurate prediction and faithful regularization can often confound the optimization of model weights.

A natural question therefore arises: is it possible to further improve the effectiveness of regularization,
potentially not only through the development of new solvers and analysis for RRM,
but also through novel mechanisms of incorporating regularization?
Although the former approach has been studied intensively,
we hypothesize and will demonstrate empirically that the latter approach can be surprisingly effective. 
Our key intuition is, now that $\Omega_i$ is specified in terms of the hidden layer output $h_i$ (which is determined by $f$),
can we directly optimize $h_i$ as opposed to indirectly through $f$?
Treating $h_i$ as ground variables and optimizing them jointly with model weights has been used by \citet{CarWan14}.
However, their motivation is on accelerating the optimization rather than improving the model.

It turns out that this idea can be conveniently implemented by leveraging the tool of proximal mapping (hence the name ProxNet),
which has been extensively used in optimization to enforce structured solutions such as sparsity \cite{ParBoy14}.
%
Given a closed convex set $C \subseteq \RR^n$ and a convex function $R: \RR^n \! \to \! \RR$ which favor certain desirable prior (\eg, $\ell_1$ norm),
the proximal mapping $\Psf_R: \RR^n \! \to \! \RR^n$ is defined as

\begin{figure}[t]
    \vspace{-0.4em}
	\centering
	\subfloat[Original two-moon dataset\label{fig:2moon_ori}] 
	{\includegraphics[width=.3\textwidth,height=0.15\textwidth]{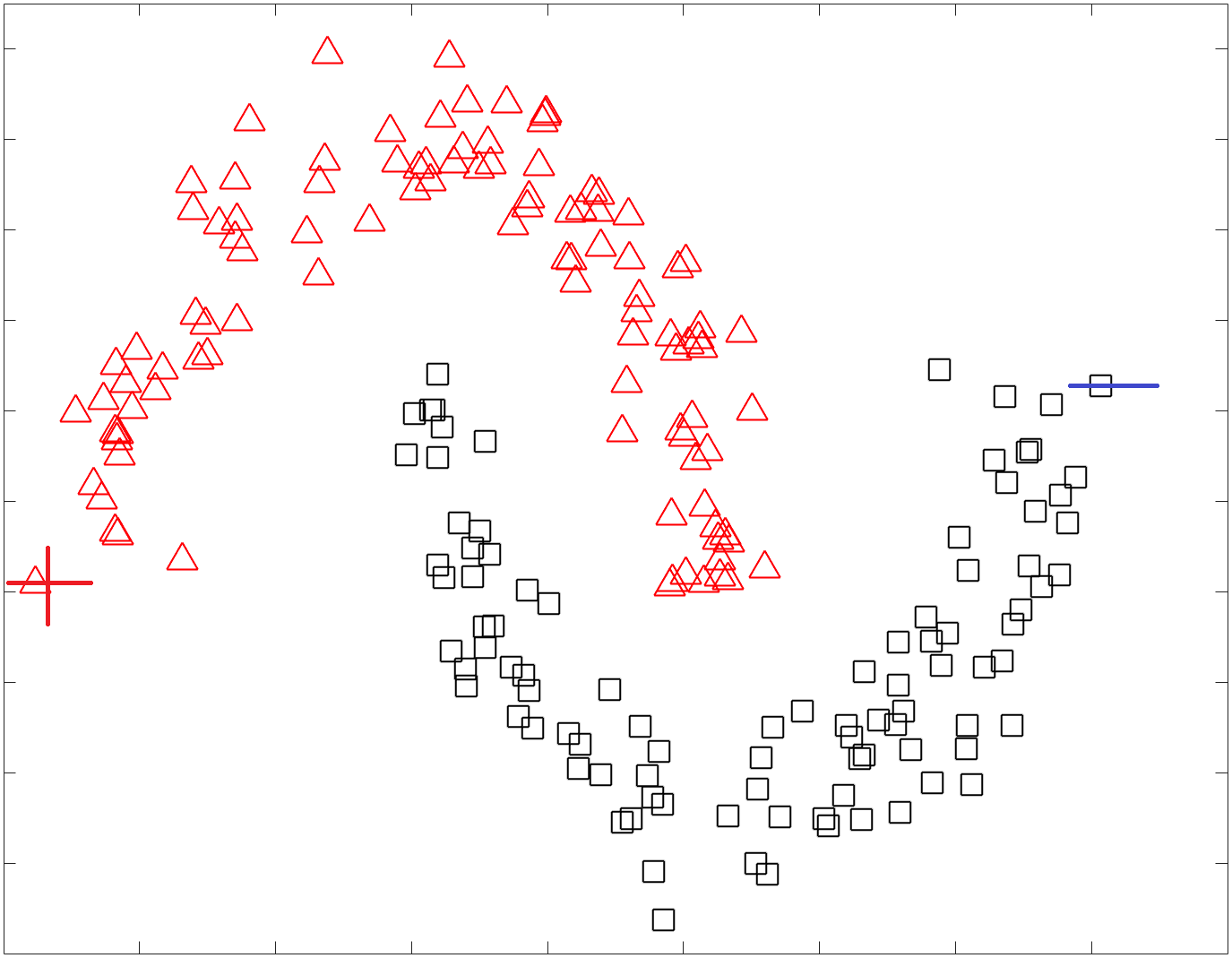}}	
	\hspace{1em}
	\subfloat[Representation after prox-map \label{fig:2moon_prox}] 
	{\includegraphics[width=.3\textwidth,height=0.15\textwidth]{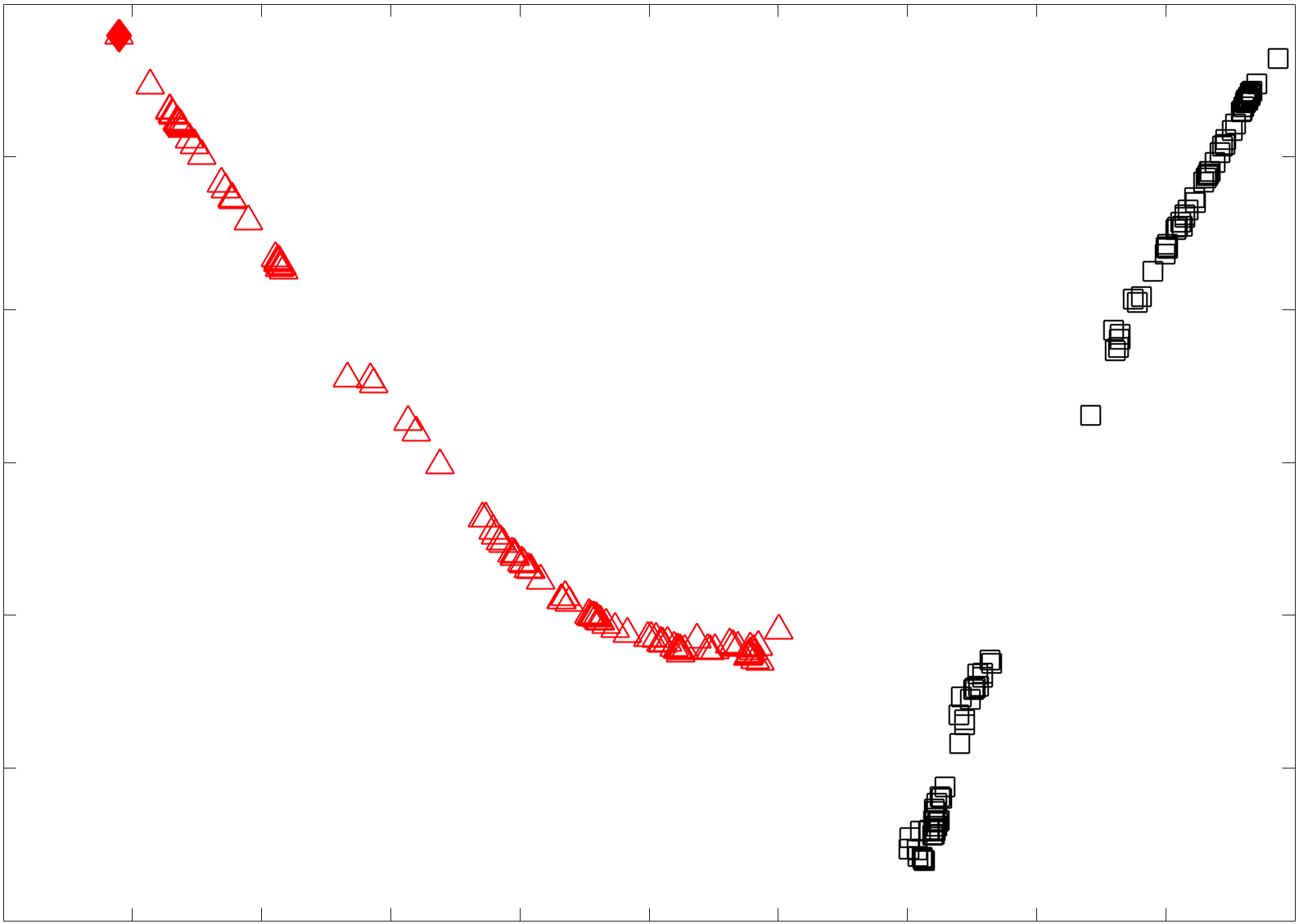}}
	\hspace{1em}
	\subfloat[Contour to {\color{red} $\blacklozenge$} after prox-map \label{fig:2moon_contour}]
	{\includegraphics[width=.3\textwidth,height=0.15\textwidth]{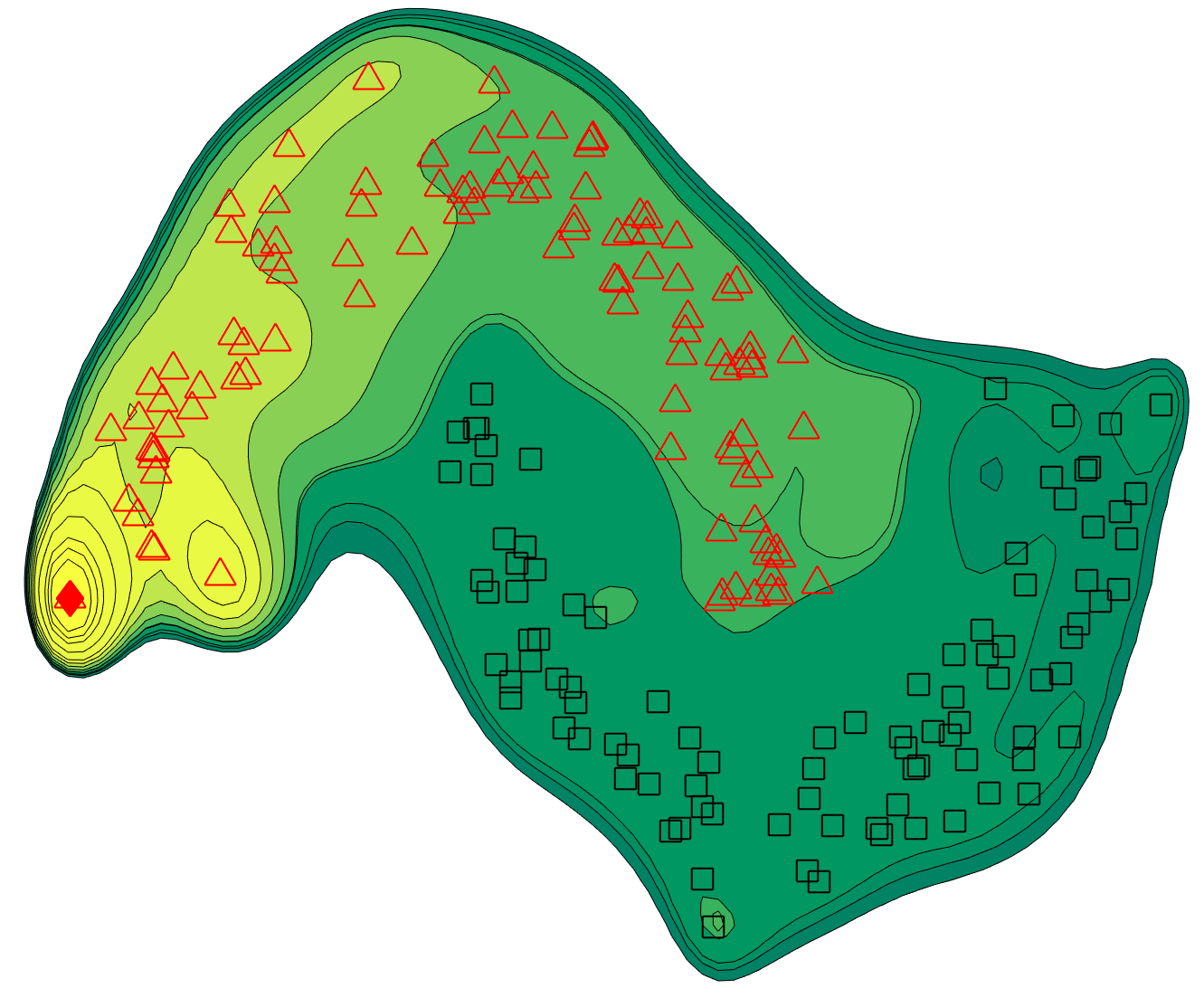}}
	\vspace{-1em}
	\caption{(a) The two-moon dataset with only two labeled examples `+' and `$-$' (left and right), but abundant unlabeled examples that reveal the inherent structure;
		(b) Representation inferred from top-2 kernel PCA based on the proximal mapping with gradient flatness and Gaussian kernel (see \S\ref{sec:shallow});
		(c) contour of distance to the leftmost point {\color{red} $\blacklozenge$}, based on the result of proximal mapping.}  
	\label{fig:twomoon}
	\vspace{-1em}
\end{figure}

\vspace{-1.8em}
\begin{align}
\label{eq:def:prox_map}
\Psf_R(x) := \argmin\nolimits_{z \in C} \{ R(z) + \smallfrac{\lambda}{2} \nbr{z - x}^2\},
\quad
\text{where the norm is } L_2.
\end{align}
%
In essence, $R$ and $C$ encourage the mapping to respect the prior encoded by $R$,
while remaining in the vicinity of $x$.
For example,
Figure \ref{fig:2moon_ori} shows the two-moon dataset with only two labeled examples and many unlabeled ones.
Figures \ref{fig:2moon_prox} and \ref{fig:2moon_contour} show the resulting representation and warped distance where $R$ accounts for the underlying manifold, 
making the classification trivial (\S \ref{sec:shallow}).

In a deep network, the proximal mapping can be inserted after any layer to turn $h_i$ into $\Psf_{\Omega_i} (h_i)$,
and backpropagate through it.
{\bf Why} does this yield a more effective implementation of regularization?
First of all,
it provides the modularity of decoupling regularization from supervised learning ---
the regularization is encapsulated \emph{within} the proximal layer that is \emph{free of weights},
and the resulting $\Psf_{\Omega_i} (h_i)$ is directly enforced to comply with the prior rather than indirectly through the optimization of weights in $f$.
This frees weight optimization from simultaneously catering to unsupervised structures and supervised performance metrics,
which plagues the conventional RRM.
Such an advantage will be confirmed in our experiments of end-to-end training that are highly efficient (\S \ref{sec:mv_comp}).

Secondly, proximal mapping can be interpreted as an intermediate step of denoising,
where $\Psf_{\Omega_i} (h_i)$ is a \emph{cleaned} version of $h_i$ that conforms to the prior.
This ensures that the downstream layers are presented with well regularized inputs,
which will presumably facilitate their own learning.
By gradually increasing $\lambda$, 
such a manual morphing can be annealed,
allowing the upstream layers (\eg, feature extractors) to approach weight values that by themselves produce well-regularized $h_i$.
ProxNet is also readily connected with meta-learning (\S \ref{sec:app_meta_learning})
because of the bi-level optimization setup,
where the proximal layer plays a similar role to base-learners.

Finally, $\Psf_{\Omega_i}(h_i)$ can be carried out on a mini-batch $\Bcal$,
where $R$ is defined on a set $\{h_i(x)\}_{x \in \Bcal}$.
It also extends flexibly to regularizers that account for multiple layers,
\eg, invariance of $h_{i}$ to $h_{i-1}$.

This paper will first review the existing building blocks of deep networks through the lens of proximal mapping 
(\S \ref{sec:preliminaries}),
and then unravel its non-trivial connections with regularization when the latter is quadratic (\eg, manifold smoothness) or non-quadratic (\eg, dropout).
Afterwards, two novel ProxNets will be introduced that achieve robust recurrent modeling (\S \ref{sec:lstm}) and multiview learning (\S \ref{sec:multiview}).
Extensive experiments show that ProxNet outperforms state-of-the-art prediction models (\S \ref{sec:experiment}).

\vspace{-0.7em}
\paragraph{Related Work}

ProxNet instantiates the differentiable optimization framework laid by OptNet \cite{AmoKol17,Agrawaletal19} along with \cite{Domke10,Domke12,BelYanMac17,BayPeaRadetal17,MetPooPfaetal17,BerHenToretal19,LeeMajRavetal19,AmoRodSacetal18,GooMirCouetal13,GouFerCheetal16}, 
which provides recipes for differentiating through an optimization layer.
In contrast, our focus is not on optimization, but on using ProxNet to model the priors in the data,
which typically involves an (inner) unsupervised learning task such as CCA.
More detailed discussions on the relationship between ProxNet and OptNet or related works are available in Appendix \ref{sec:compare_optnet}.

Another proximal-like operator was found in ``sparsemap'' operations \citep{NicBlo17,NicMarBloCar18,MarAst16}.
However, they target a different application of incorporating structured sparsity in attention weights for a \textit{single} instance, 
rather than at a mini-batch level where ProxNet is applied for multiview learning.

\vspace{-0.8em}
\section{Proximal Mapping as a Primitive Construct in Deep Networks}
\label{sec:preliminaries}
\vspace{-0.5em}


Proximal mapping is highly general, encompassing most primitive operations in deep learning \cite{ParBoy14,ComPes18}.
For example, any activation function $\sigma$ with $\sigma'(x) \in (0,1]$ (\eg, sigmoid) is indeed a proximal map with $C = \RR^n$ and $R(x) = \int \sigma^{-1} (x) \rmd x - \frac{1}{2} x^2$,
which is convex.
The ReLU and hard tanh activations can be recovered by $R = 0$, 
with $C = [0, \infty)$ and $C = [-1,1]$, respectively.
Soft-max transfer $\RR^n \ni x \mapsto (e^{x_1}, \ldots, e^{x_n})^\top / \sum_i e^{x_i}$ corresponds to
$C = \{x \in \RR_+^n : \one^\top x = 1\}$ and
$R(x) = \sum_i x_i \log x_i - \frac{1}{2} x_i^2$, which are convex.
Batch normalization maps $x \in \RR^n$ to $(x-\mu \one)/\sigma$,
where $\one$ is a vector of all ones,
and $\mu$ and $\sigma$ are the mean and standard deviation of the elements in $x$, respectively.
This mapping can be recovered by $R = 0$ and $C = \{x: \nbr{x} = \sqrt{n}, \one^\top x = 0\}$.
Although $C$ is not convex, this $\Psf_R(x)$ must be a singleton for $x \neq 0$.
In general, $R$ and $C$ can be nonconvex making $\Psf_R(z)$ set-valued,
and we only need differentiation at one element
\cite{HarSag09,BerThi04,PolRoc96}.


\vspace{-0.8em}
\paragraph{Kernelization.}

Proximal mapping can be trivially extended to reproducing kernel Hilbert spaces (RKHSs),
allowing non-vectorial data to be encoded \cite{LafCledAl19}
and invariances to be hard wired \cite{MaiKonHaretal14,BieMai17}.
%
%
Assume an RKHS $\Hcal$ employs a kernel $k:$ $\Xcal \times \Xcal \to \RR$ with an inner product $\inner{\cdot}{\cdot}_\Hcal$.
Given a convex functional $R: \Hcal \to \RR$,
a proximal map $\Psf_R : \Hcal \to \Hcal$ can be defined in exactly the same form as \eqref{eq:def:prox_map}, with $L_2$ norm replaced by RKHS norm.
\vspace{-0.5em}
\section{Connecting Proximal Mapping to RRM on Shallow Models}
\label{sec:shallow}
\vspace{-0.4em}

We first illustrate the connection between RRM and proximal mapping.
To focus on the core idea, we use shallow models with no hidden layer.
Letting $k_x := k(x,\cdot)$ be the kernel representer of $x$ and
$R$ be the regularizer encoding preference on $f$,
we can write the two formulations as follows:
\begin{align}
    {\color{blue}\text{P1: }}
    \min\nolimits_{f \in \Hcal} \EE_{x \sim \ptil} &[\ell(\inner{f}{{\color{blue}k_x}}_\Hcal)] + R(f)
    \quad
    \text{v.s.}
    \quad
    {\color{red}\text{P2: }}
    \min\nolimits_{h \in \Hcal} \EE_{x \sim \ptil} [\ell(\inner{h}{{\color{red}c_x}}_\Hcal)] + \lambda^2 \nbr{h}^2_\Hcal, \!\! \\
    \label{eq:def_prox_kernel}
    \where 
    {\color{red}c_x} &:= \Psf_R({\color{blue}k_x}) = \argmax\nolimits_{g \in \Hcal} 
    \cbr{\smallfrac{\lambda}{2} \nbr{g - {\color{blue}k_x}}_\Hcal^2 + R(g)}.
\end{align}

\paragraph{i) $R(f)$ is a positive semi-definite (PSD) quadratic.}
Examples of this simplest case include graph Lapalacian
$R_{l}(f) := \! \sum_{ij} w_{ij} (f(x_i)  -  f(x_j))^2$
and gradient penalty
$R_{g}(f) := \! \sum_{i} \nbr{\grad f(x_i)}^2$.
They both enforce smoothness on a data manifold.
Since the gradient operator $\grad R: f \mapsto \grad R(f)$ is linear,
we denote its eigenvalues and eigenfunctions as $\{\mu_i, \phi_i\}$.
Further, taking derivative of $g$ in \eqref{eq:def_prox_kernel},
we derive a closed-form for the proximal map as 
$c_x = \lambda (\lambda I + \grad R)^{-1} k_x$,
where $I$ is the identity operator.
The contour in Figure \ref{fig:2moon_contour} was plotted exactly by using the pairwise distance $\nbr{c_x - c_{x'}}_\Hcal$,
based on which the new data representation in Figure \ref{fig:2moon_prox} was extracted using the top-2 principal components.

To connect P1 and P2, let $h \! = \! \lambda^{-1} (\lambda I + \grad R) f$.
Then $\inner{f}{k_x}_\Hcal \! = \! \inner{h}{c_x}_\Hcal$ (\ie, same prediction)
and
\begin{align}
    R(f) = \smallfrac{1}{2} \sum\nolimits_i \mu_i \inner{f}{\phi_i}_\Hcal^2,
    \qquad
    \text{and}
    \qquad
    \lambda^2 \nbr{h}_\Hcal^2 = \sum\nolimits_i  (\lambda + \mu_i)^2  \inner{f}{\phi_i}_\Hcal^2.
\end{align}
This reveals that P1 and P2 are connected through a \emph{monotonic} spectral transformation.
When $\lambda$ is small, it simply squares the eigenvalues,
which leads to little difference in learning as we observed in experiment.
Moreover, there is a similar connection between $c_x$ and the kernel representer of a new RKHS, 
which warps the original RKHS norm into $\nbr{f}_\Hcal^2 + R(f)$ \citep{MaGanZha19}.
See details in Appendix \ref{sec:app_warping}.

\paragraph{ii) General $R$.}

When $R$ is not quadratic,
the linear relationship between $c_x$ and $k_x$ no longer exists.
However, some relaxed connection between P1 and P2 is still available,
and we will demonstrate it on dropout training.
As discovered by \citet{WagWanLia13,WanMan13}, 
dropout on input features in a single-layer network leads to an adaptive regularizer on a linear discriminant 
$x \mapsto \beta^\top x$ (derivation is in Appendix \ref{sec:app_dropout}):
\begin{align}
\label{eq:reg_dropout}
	R_{\ptil}(\beta) =
	\sum\nolimits_i \EE_{x \sim \ptil} [p_x (1 - p_x) x_i^2] \cdot \beta_i^2,
	\where
	p_x := \sigma(x^\top \beta) := (1 + \exp(- x^\top \beta))^{-1}.
\end{align}

Here $R_\ptil$ penalizes $\beta_i$ more mildly if $x_i$ is generally small.
This allows rare but discriminative features to receive higher weights,
which is useful in text data.
Now to connect P1 and P2,
we simplify the computation by using the proximal map of $R_{\delta_x}$ instead of $R_\ptil$,
where $\delta_x$ is the Dirac distribution at $x$:
\begin{align}
\label{eq:pm_dropout}
    c_x&:= \Psf_{R_{\delta_x}}(x) = \argmin\nolimits_{c} \ \cbr{\smallfrac{1}{2} \sum\nolimits_i p_x(1-p_x) x_i^2 c_i^2 + \smallfrac{\lambda}{2} \nbr{c - x}^2},
    \text{ where } p_x = \sigma(x^\top c).
\end{align}
Since $p_x$ depends only on $x^\top c$,
we first fix $x^\top c$ to $s$, 
hence $p_x (1-p_x) = \alpha_s := \frac{2}{2+e^s + e^{-s}}$.
Enforcing $x^\top c = s$ by a Lagrange multiplier $\mu$, 
$c_i$'s are decoupled, 
allowing them to be optimized analytically:
\begin{align}
    (c_x)_i &= (\lambda + \mu) x_i (\lambda + \alpha_s x_i^2)^{-1},
\ \ \ \where
\mu \text{ is such that } x^\top c_x = s.
\end{align}
\newpage

Finally \eqref{eq:pm_dropout} can be optimized through a 1-D line search on $s$.
Letting $h_i = \beta_i \frac{\lambda + \alpha_s x_i^2}{\lambda + \mu}$,
we have $\beta^\top x = h^\top c_x$ (same predictions) and 
$
    \nbr{h}^2 = (\lambda+\mu)^{-2} \sum_i (\lambda + p_x (1-p_x) x_i^2)^2 \beta_i^2,
$
which resembles $R_\ptil$ in \eqref{eq:reg_dropout} especially when $\lambda$ is small.
However, since $\frac{h_i}{\beta_i}$ depends on $x$,
this reformulation meets with difficulty when extended to the whole dataset $\ptil$.
We emphasize that our aim here is to shed light on the connection between regularization and proximal mapping;
we do \emph{not} intend to establish their exact equivalence.
Simulations in Appendix \ref{sec:app_dropout} show that P1 and P2 deliver very similar predictions.

\vspace{-0.8em}
\paragraph{Application to multiple layers.}
It is straightforward to apply proximal mapping to any hidden layer of interest and for multiple times.
A similar warping trick  was introduced in \citet{MaGanZha19} to invariantize  convolutional kernel descriptors \citep{MaiKonHaretal14,Mairal16}.
However it was restricted to linear invariances.
Proximal mapping, instead, lifts this restriction by accommodating nonlinear invariances such as total variation.

\vspace{-0.4em}
\section{Proximal Mapping for Robust Learning in Recurrent Neural Nets}
\label{sec:lstm}
\vspace{-0.25em}

Our first novel instance of ProxNet tries to invariantize LSTM to perturbations on inputs $x_t$.
Virtual adversarial training has been proposed in this context as an unsupervised regularizer \citep{MiyDaiGoo17},
where the underlying prior postulates that such robustness can benefit the prediction accuracy.
The resilience under real attack, however, is \textit{not} the main concern in \citep{MiyDaiGoo17}.
We will demonstrate empirically that this prior can be more effectively implemented by ProxNet,
leading to improved prediction performance.

The dynamics of hidden states $c_t$ in an LSTM can be represented by
$c_t = f(c_{t-1}, h_{t-1}, x_t)$,
with outputs $h_t$ updated by $h_t = g(c_{t-1}, h_{t-1}, x_t)$.
We aim to encourage that the hidden state $c_t$ stays invariant, 
when each $x_t$ is perturbed by $\delta_t$ whose norm is bounded by $\delta$.
To this end, we introduce an intermediate step 
$s_t = s_t(c_{t-1}, h_{t-1}, x_t)$ that computes the original hidden state,
and then apply proximal mapping so that the next state $c_t$ remains close to $s_t$,
while also moving towards the \emph{null} \emph{space} of the variation of $s_t$ under the perturbations on $x_t$.
Formally, using first-order approximation,

\vspace{-1.7em}
\begin{align*}
\nonumber
c_t :&= \argmin_c \smallfrac{\lambda}{2} \nbr{c - s_t}^2 
	+ \smallfrac{1}{2}\max\nolimits_{\nbr{\delta_t} \le \delta} 
	\langle c, s_t(c_{t-1},h_{t-1}, x_t) - s_t(c_{t-1}, h_{t-1}, x_t + \delta_t) \rangle^2 \\
&\approx \argmin_c \lambda \nbr{c \! - \! s_t}^2 \!\! + \!\!
    \max\nolimits_{\nbr{\delta_t} \le \delta} 
	\langle c, \smallfrac{\partial}{\partial x_t} s_t(c_{t-1}, h_{t-1}, x_t) \delta_t \rangle^2 \\
&= \argmin_c \lambda \nbr{c - s_t}^2 
+ \delta^2\nbr{c^\top G_t}_*^2, \quad \text{where }\quad G_t := \smallfrac{\partial}{\partial x_t} s_t(c_{t-1}, h_{t-1}, x_t)
\end{align*}

\begin{figure*}[t!]
	\vspace{-0.4em}
	\centering
		\begin{minipage}[b]{0.35\textwidth}
 	\includegraphics[width=0.99\linewidth]{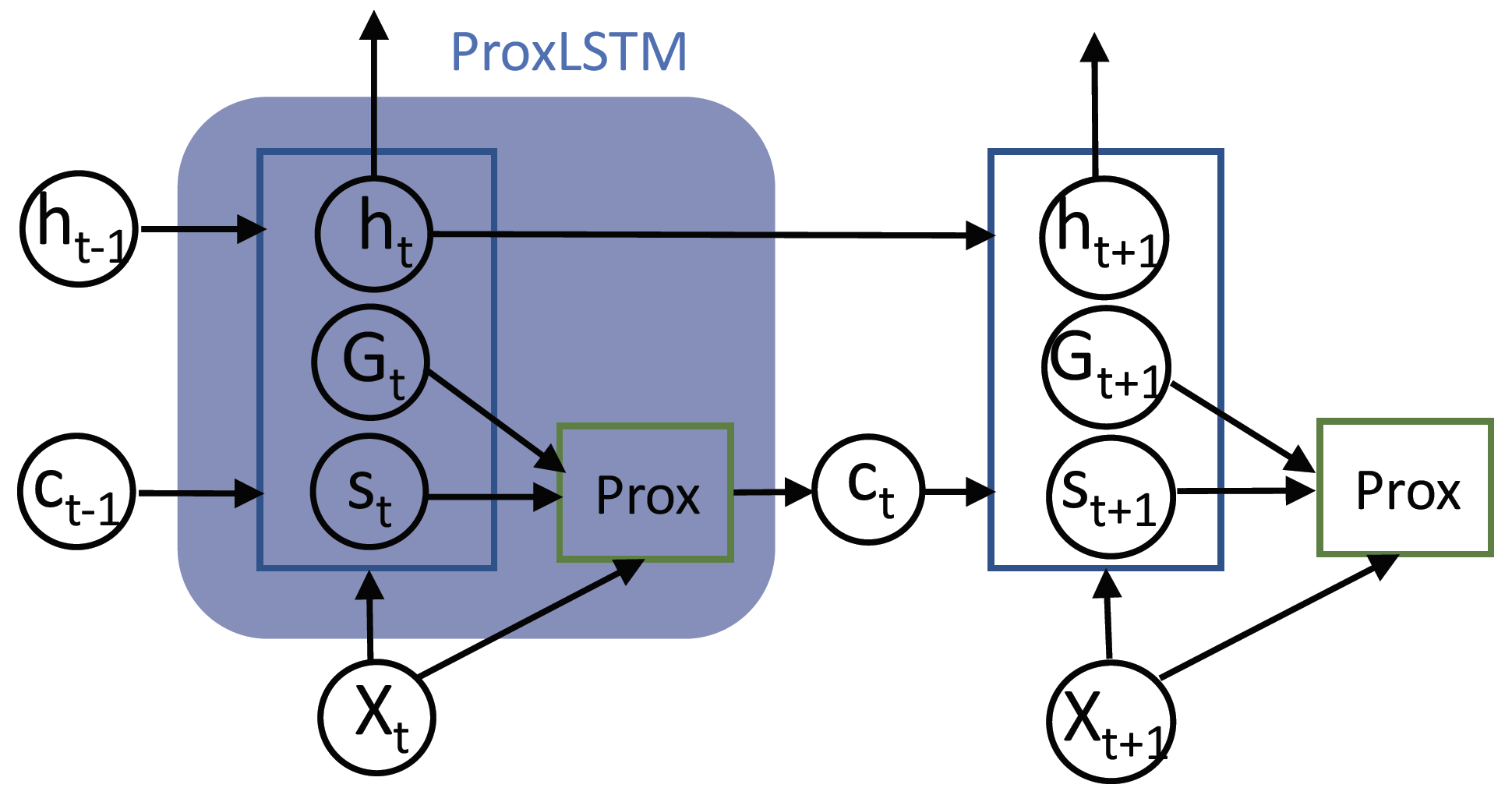}
	\vspace{-1em}
	\caption{A proximal LSTM layer}
	\label{fig:proxLSTM}
	\end{minipage}
	~
	\begin{minipage}[b]{0.63\textwidth}
	\includegraphics[width=0.99\linewidth]{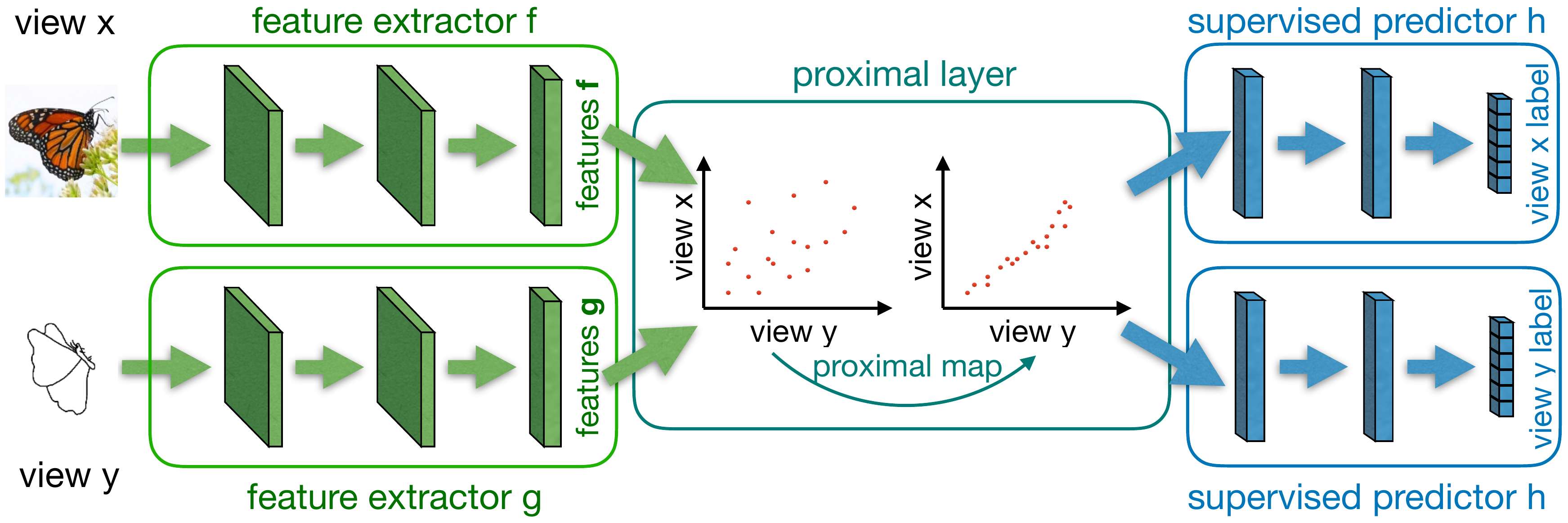}
	\vspace{-0.5em}
	\caption{ProxNet for multiview learning with proximal CCA}
	\label{fig:multiview}
	\end{minipage}
	\vspace{-1.9em}
\end{figure*}

\vspace{-0.9em}
and $\nbr{\cdot}_*$ is the dual norm.
The diagram is shown in Figure \ref{fig:proxLSTM}.
Using the $L_2$ norm, 
a closed-form solution for $c_t$ is
$(I + \lambda^{-1} \delta^2 G_t G_t^\top)^{-1} s_t$,
and BP can be reduced to second-order derivatives (\S \ref{sec:bp_lstm}).
A key advantage of this framework is the generality and ease in inserting proximal layers into the framework ---
simply invoke the second-order derivatives of the underlying (gated) units as a black box. 
We will refer to this model as ProxLSTM.

\vspace{-0.4em}
\section{ProxNet for Multiview Learning}
\label{sec:multiview}
\vspace{-0.3em}

While proximal mapping is applied on each individual data point in ProxLSTM,
it can indeed be applied in mini-batches,
and we next demonstrate its application in multiview learning with sequential structures.
Here each instance exhibits a pair of views:
$\{(x_i, y_i)\}_{i=1}^n$,
and is associated with a label $c_i$.
In the deep canonical correlation analysis model \citep[DCCA,][]{AndAroBiletal13},
the $x$-view is passed through a multi-layer neural network or kernel machine,
leading to a hidden representation $f(x_i)$.
Similarly the $y$-view is transformed into $g(y_i)$.
CCA aims to maximize the correlation of these two views after projecting into a common $k$-dimensional subspace, through $\{u_i\}_{i=1}^k$ and $\{v_i\}_{i=1}^k$ respectively.
Denoting $X = (f(x_1), \ldots, f(x_n)) H$ and $Y = (g(y_1), \ldots, g(y_n)) H$
where $H = I - \frac{1}{n} \one \one^\top$ is the centering matrix,
CCA finds $U = (u_1, \ldots, u_k)$ and $V = (v_1, \ldots, v_k)$ that maximize the correlation:
\vspace{-0.25em}
\begin{align}
\label{eq:obj_cca}
	\min\nolimits_{U, V} -\tr(U^\top X Y^\top V), 
	\ \ \ \text{s.t.} \ \ U^\top X X^\top U = I, \  V^\top Y Y^\top V = I, \
	 u_i^\top X Y^\top v_j = 0, \ \forall i \neq j.\!
\end{align}
\newpage

Denote the optimal objective value as $L(X, Y)$.
DCCA directly optimizes it with respect to the parameters in $f$ and $g$,
while DCCA autoencoder \citep[DCCAE,][]{WanAroLivetal15} further reconstructs the input.
They both use the result to initialize a finer tuning of $f$ and $g$, 
in conjunction with subsequent layers $h$ for a supervised target $c_i$.
We aim to improve this two-stage process with an end-to-end approach based on proximal mapping, which can be written  as
$\min_{f, g, h} \sum_i \ell(h(p_i, q_i), c_i)$
where $\{(p_i, q_i)\}_{i=1}^n$ is from
\begin{align}
\label{eq:prox_multiview}
    \Psf_L(X, Y) = \argmin\nolimits_{P, Q} \ {\smallfrac{\lambda}{2n} \nbr{P - X}_F^2 +
	\smallfrac{\lambda}{2n} \nbr{Q - Y}_F^2 + L(P, Q)}.
\end{align}
Here $\nbr{\cdot}_F$ stands for the Frobenius norm,
$P = (p_1, \ldots, p_n)$,
and $Q = (q_1, \ldots, q_n)$.
Clearly, \eqref{eq:prox_multiview} has applied proximal mapping to a \emph{mini-batch},
and we will show how to save computational cost, especially at test time.
The entire framework is illustrated in Figure \ref{fig:multiview}.


\vspace{-0.2em}
\subsection{Backpropagation and computational cost}
\label{sec:mv_comp}


Although efficient closed-form solution is available for the CCA objective in \eqref{eq:obj_cca},
none exists for the proximal mapping in \eqref{eq:prox_multiview}.
However, it is natural to take advantage of this closed-form solution.
In particular, assuming $f(x_i)$ and $g(y_i)$ have the same dimensionality,
\citet{AndAroBiletal13} showed that $L(X,Y) = -\sum_{i=1}^k \sigma_i(T)$,
where $\sigma_i$ is the $i$-th largest singular value,
and
%
\begin{align*}
  T(X, Y) = (X X^\top + \epsilon I)^{-1/2} (X Y^\top) (Y Y^\top + \epsilon I)^{-1/2}.
\end{align*}
%
Here $\epsilon > 0$ is a small stabilizing constant.
Then 
%
%
\eqref{eq:prox_multiview} can be solved by gradient descent or L-BFGS.
The gradient of $\sum_{i=1}^k \sigma_i(T(P,Q))$ is available from \citet{AndAroBiletal13},
which relies on SVD.
Although SVD appears expensive,
fortunately, the cost of computing $T$ and SVD is low in practice because
i) the dimensions of $f$ and $g$ are low in practice (10 in our experiment and DCCA),
and ii) the mini-batch size does not need to be large.
In our experiment, increasing mini-batch size beyond 100 did not significantly improve the performance.
Extension to more than two views is relegated to Appendix \ref{sec:app_multiview}.

Backpropagation through the proximal mapping in \eqref{eq:prox_multiview} requires that given $\frac{\partial J}{\partial P}$ and $\frac{\partial J}{ \partial Q}$ where $J$ is the ultimate objective value,
compute $\frac{\partial J}{\partial X}$ and $\frac{\partial J}{ \partial Y}$.
The most general solution has been provided by OptNet \citep{AmoKol17,Agrawaletal19}, 
but the structure of our problem admits a simpler solution from \citet{Domke10}.
\begin{align*}
    \rbr{\smallfrac{\partial J}{ \partial X}, \smallfrac{\partial J}{ \partial Y}} 
 \approx \smallfrac{1}{\epsilon}  \big(\Psf_L(X  +  \epsilon \smallfrac{\partial J}{\partial P}, Y  +  \epsilon \smallfrac{\partial J}{ \partial Q})  -  \Psf_L(X, Y)\big),
 \qquad 0 < \epsilon \ll 1.
\end{align*}
%
To reduce the test time complexity for ProxNet,
we draw a key insight that if the feature extractor preceding the proximal mapping is well trained so that the latent representation of the two views is highly correlated, then the proximal layer may improve performance only marginally.

Therefore, we can take advantage of proximal mapping during training, 
while gradually fade it out at the fine tuning stage. 
Towards this end, the weight $\lambda$ that controls the trade off between correlation and displacement can be increased as training proceeds.
More specifically, we set in experiment $\lambda_t = (1+kt) \alpha_0$ at epoch $t$, where $\alpha_0$ and $k$ are hyperparameters.
As a result, test time predictions can be made very efficiently by dispensing with proximal mapping or mini-batch.



\vspace{-0.6em}
\paragraph{Extension to recurrent networks.}

ProxNet can be readily extended to structured data.
As illustrated in Figure \ref{fig:ProxNet_seq},
an RNN can be used as a feature extractor,
and the hidden units of the two views are fed into proximal mapping.
In the simplest formulation, all hidden units are treated independently,
leaving the sequential structure to LSTM.
A more refined approach can retain or even add spatio-temporal structures inside the proximal mapping,
\eg, total variation and permutation invariance.
We will use the simplest form in our experiment for speech recognition.


\section{Experimental Results}
\label{sec:experiment}
\vspace{-0.2em}

We evaluated the empirical performance of ProxNet for multiview learning on supervised learning (two tasks) and unsupervised learning (crosslingual word embedding). 
ProxLSTM was evaluated on sequence classification.
We used the Ray Tune library to select the hyper-parameters for all baseline methods \citep{Liawetal18}.
Details on data preprocessing, experiment setting, optimization, and additional results are given in Appendix \ref{sec:app_exp}.
Here we highlight the major results and experiment setup.

\vspace{-0.7em}
\paragraph{Baselines.}

For the three multiview tasks, 
we will demonstrate the effectiveness of ProxNet by comparing with state-of-the-art methods including \textbf{DCCA} and \textbf{DCCAE}.
Neither DCCA nor DCCAE is end-to-end training, and a classifier was trained on their hidden code.
As a basic competitor, we also considered a \textbf{Vanilla} method, which trained a network for each view independently. 
\newpage

\begin{minipage}[t!]{\textwidth}
	\begin{minipage}[t!]{0.48\textwidth}
	    \vspace{1.2em}
		\centering
    	\captionof{figure}{ProxNet for multiview sequential data}
    	\vspace{-0.4em}
    	\includegraphics[height=3cm, width=\linewidth]{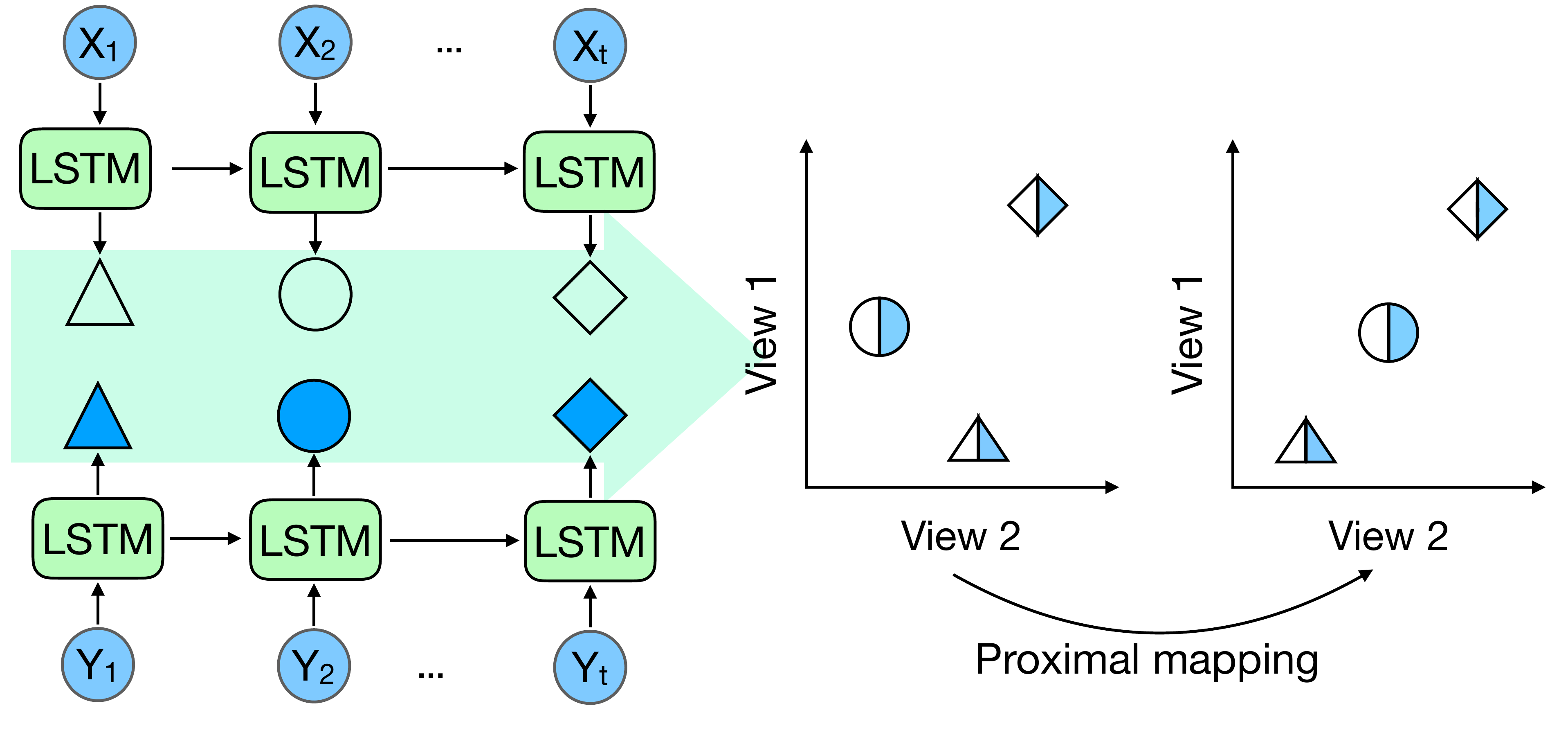}
    	\label{fig:ProxNet_seq}
	\end{minipage}
	\hfill
	\begin{minipage}[t!]{0.52\textwidth}
		\captionof{table}{Average test error ($\%$) on Sketchy}
    	\label{tab:sketchy}
    	\vspace{-0.4em}
    	\centering
    	\setlength\tabcolsep{2pt}
    	\begin{tabular}{@{}rccccc@{}}
    		\toprule
    		\#class &  20 & 50 &  100 &  125 \\
    		\midrule
    		Vanilla & 18.7  {\scriptsize $\pm$ 1.1} & 24.8  {\scriptsize $\pm$ 0.9} 
    		        & 30.9  {\scriptsize $\pm$ 0.5} & 31.8  {\scriptsize $\pm$ 0.5} \\
    		DCCA    & 16.9  {\scriptsize $\pm$ 0.5} & 22.2  {\scriptsize $\pm$ 0.4} 
    				& 28.7  {\scriptsize $\pm$ 0.4} & 29.8  {\scriptsize $\pm$ 0.4} \\
    		DCCAE   & 16.6  {\scriptsize $\pm$ 0.3} & 22.1  {\scriptsize $\pm$ 0.3} 
    				& 29.2  {\scriptsize $\pm$ 0.5} & 30.4  {\scriptsize $\pm$ 0.6} \\
    		RRM & 15.2  {\scriptsize $\pm$ 0.6} & 20.1  {\scriptsize $\pm$ 0.4} 
    		        & 26.8  {\scriptsize $\pm$ 0.5} & 28.1  {\scriptsize $\pm$ 0.4} \\
    		\midrule
    		ProxNet  & \textbf{13.7}  {\scriptsize $\pm$ 0.3} & \textbf{17.9}  {\scriptsize $\pm$ 0.5} 
    				 & \textbf{20.2}  {\scriptsize $\pm$ 0.3} & \textbf{22.0}  {\scriptsize $\pm$ 0.4} \\
    		\bottomrule
    	\end{tabular}
	\end{minipage}
	\vspace{-0.4em}
\end{minipage}

Our key competitor is RRM,
which motivated ProxNet in the introduction section.
Specifically, it moves the regularizer $L(X,Y)$ defined in \eqref{eq:obj_cca} from inside the proximal mapping to the overall objective as in \eqref{eq:obj_RRM},
promoting the correlation between the two views' hidden representation through the network weights.
At test time, ProxNet, RRM, and Vanilla all made predictions by averaging the logits from both views.
This consistently outperformed concatenating the logits of the two views.

\vspace{-0.3em}
\subsection{Multiview supervised learning 1: image recognition with sketch and photo}
\label{sec:sketch_exp}



\vspace{-0.7em}
\paragraph{Dataset.} 
We first evaluated ProxNet on a large-scale sketch-photo paired database  ``Sketchy'' \citep{sketchy2016}.
It consists of 12,500 photos and 75,471 hand-drawn sketches of objects from 125 classes. 
Each instance is a pair of sketch and photo representing the same natural image,
both in color and sized $256\times 256$.
To demonstrate the robustness of our method, we varied the number of classes over $\{20, 50, 100, 125\}$ by sampling a subset from the original dataset.
For each class, there are 100 sketch-photo pairs.
We randomly sampled 80 pairs from each class to form the training set, 
and then used the remaining 20 pairs for testing.

\vspace{-0.7em}
\paragraph{Implementation details.}
Unless otherwise specified, our implementations were based on PyTorch and all training was conducted on a NVIDIA GeForce 2080 Ti GPU.
All methods were trained using ResNet-18 as the feature extractor.
In ProxNet, the proximal layer has input and output dimension $d = 20$,
followed by three fully-connected layers of 512 hidden units with sigmoid activations.
The final output layer has multiple softmax units, 
each corresponding to an output class.
ProxNet was trained by Adam with a weight decay of $0.0001$ and a learning rate of $0.001$,
with the latter divided by 10 after 200 epochs.
The mini-batch size was $100$, which, in conjunction with the low dimensionality of proximal layer ($d=20$),
allows the SVD to be solved instantaneously.
At training time, we employed an adaptive trade-off parameter $\lambda$,
which is defined in \eqref{eq:prox_multiview}.
We set the hyper-parameters $k = 0.5$ and $\alpha_0 = 0.1$.
All experiments were run five times to produce mean and standard deviation.

\vspace{-0.7em}
\paragraph{Results.}
As shown in Table \ref{tab:sketchy}, 
ProxNet delivers significantly lower test error than all other baselines.
Interestingly, the improvement becomes more significant with the increasing number of classes.
Vanilla performs the worst,
and RRM outperforms DCCA and DCCAE thanks to end-to-end training.

\vspace{-0.5em}
\subsection{Multiview supervised learning 2: audio-visual speech recognition}

Our second task aims to learn features and classifiers for speaker-independent phonetic recognition.

\vspace{-0.8em}
\paragraph{Dataset.}
We used the Wisconsin X-ray Micro-Beam Database (XRMB) corpus which consists of simultaneously recorded \emph{speech} and \emph{articulatory} measurements from 47 American English speakers and 2357 utterances \citep{Westbury94}. 
The first view is acoustic features comprising 39D mel frequency cepstral coefficients (MFCCs) and their first and second derivatives,
and the second view is articulatory features made up of 16D horizontal/vertical displacement of 8 pellets attached to several parts of the vocal tract.
Also available is the phonetic labels for classification.
To simulate the real-life scenarios and to improve the model's robustness to noise, 
we corrupted the acoustic features of a given speaker by mixing with 
$\{0.2, 0.5, 0.8\}$ level of another randomly picked speaker's acoustic features. 
The whole dataset was partitioned into 35 speakers for training and 12 speakers for testing.


\vspace{-0.7em}
\paragraph{Implementation details.}
To incorporate context information, 
\citet{WanAroLivetal15} concatenated the inputs over a window sized $W$ centered at each frame, giving $39\times W$ and $16\times W$ feature dimensions for each of the two views respectively. 
Although this delicately constructed input freed the encoder/feature extractor from considering the time dependency within frames,
we prefer a more refined modeling of the sequential structure,
and therefore adopted, for all methods under comparison, a 2-layer LSTM with hidden layers of $256$ units, 
followed by a fully-connected layer which projects the outputs of LSTM to a $K$-dimensional subspace, 
serving as the feature extractor. 

The supervised predictor is a fully-connected network with an output layer of 41 softmax units. 
We used the Connectionist Temporal Classification loss \citep[CTC,][]{AleSanFau06}, 
which adopts greedy search as the phone recognizer.
The dimension of subspace was tuned in $\{10, 20, 30, 50\}$, 
and the sequence length was tuned in $\{250, 500, 1000\}$ for all algorithms.
The mini-batch size was set to 32.
Although the proximal mapping here solves a larger problem than that in \S \ref{sec:sketch_exp},
we observed that a higher value of $\lambda$ was sufficient to enforce a high correlation on this dataset,
hence keeping the optimization efficient. 
In practice, we set $k = 1$ and $\alpha_0 = 0.5$.

In order to compare the effectiveness of different algorithms in information transfer without being confounded by logit averaging (logit-avg) which can achieve a similar effect, 
we studied another mode called ``acoustic''.
Here all algorithms predict on test data by only using the output layer of the acoustic view,
and at training time a loss is applied to each view based on the ground truth label.

\vspace{-1em}
\paragraph{Results.}

Table \ref{tab:pers} presents the Phone Error Rates (PERs) of all methods.
Clearly, ProxNet achieves the lowest PER among all algorithms at all levels of noise.
The margin over the runner-up (RRM) is the largest when there is no noise.
As expected, ``logit-avg'' almost always outperforms ``acoustic'',
because the articulatory features are clean,
supplying reliable predictions.
Focusing on the ``acoustic'' columns,
Vanilla cannot leverage articulatory features,
while other methods can achieve it by promoting correlations in the hidden space.
ProxNet appears most effective in this respect.

\begin{table*}[t!]
\setlength\tabcolsep{3pt}
	\caption{Mean and standard deviation of PERs on the XRMB dataset with different noise levels}
	\label{tab:pers}
	\vspace{-0.4em}
	\centering
	\begin{tabular}{@{}rcc|cc|cc|cc@{}}
		\toprule
		& \multicolumn{2}{c|}{noise level = 0\%} & \multicolumn{2}{c|}{noise level = 20\%} & \multicolumn{2}{c|}{noise level = 50\%} & \multicolumn{2}{c}{noise level = 80\%} \\
		\cmidrule{2-3} \cmidrule{4-5} \cmidrule{6-7} \cmidrule{8-9}
		& acoustic & logit-avg & acoustic & logit-avg & acoustic & logit-avg & acoustic & logit-avg\\
		\midrule
		Vanilla & 17.9 {\scriptsize $\pm$ 1.0} & 17.1 {\scriptsize $\pm$ 0.6}
		        & 19.3 {\scriptsize $\pm$ 0.8} & 19.1 {\scriptsize $\pm$ 1.2} 
		        & 27.7 {\scriptsize $\pm$ 1.1} & 21.4 {\scriptsize $\pm$ 0.8} 
		        & 45.1 {\scriptsize $\pm$ 1.0} & 24.4 {\scriptsize $\pm$ 1.0}\\
		DCCA    & 17.3 {\scriptsize $\pm$ 0.3} & 16.3 {\scriptsize $\pm$ 0.5}
		        & 18.8 {\scriptsize $\pm$ 0.3} & 16.3 {\scriptsize $\pm$ 0.6} 
		        & 26.0 {\scriptsize $\pm$ 0.3} & 23.6 {\scriptsize $\pm$ 0.5} 
		        & 45.1 {\scriptsize $\pm$ 0.9} & 34.9 {\scriptsize $\pm$ 0.9}\\
		DCCAE   & 15.5 {\scriptsize $\pm$ 0.2} & 15.3 {\scriptsize $\pm$ 0.4}
		        & 16.7 {\scriptsize $\pm$ 0.3} & 15.9 {\scriptsize $\pm$ 0.4} 
		        & 23.6 {\scriptsize $\pm$ 0.3} & 21.8 {\scriptsize $\pm$ 0.7} 
		        & 43.9 {\scriptsize $\pm$ 0.7} & 34.8 {\scriptsize $\pm$ 0.7}\\
		RRM  & 16.1 {\scriptsize $\pm$ 0.5} & 15.0 {\scriptsize $\pm$ 0.3}
		             & 16.6 {\scriptsize $\pm$ 0.7} & 16.9 {\scriptsize $\pm$ 0.5} 
		             & \textbf{22.3} {\scriptsize $\pm$ 0.8} & 21.6 {\scriptsize $\pm$ 0.6} 
		             & 40.7 {\scriptsize $\pm$ 0.7} & 23.9 {\scriptsize $\pm$ 0.3}\\
		\midrule
		ProxNet & \textbf{12.9} {\scriptsize $\pm$ 0.4} & \textbf{10.5} {\scriptsize $\pm$ 0.4}
				& \textbf{15.3} {\scriptsize $\pm$ 0.4} & \textbf{11.2} {\scriptsize $\pm$ 0.3} 
				& \textbf{21.6} {\scriptsize $\pm$ 0.5} & \textbf{16.6} {\scriptsize $\pm$ 0.3} 
				& \textbf{39.3} {\scriptsize $\pm$ 0.3} & \textbf{20.1} {\scriptsize $\pm$ 0.5}\\
		\bottomrule
	\end{tabular}
	\vspace{-0.7em}
\end{table*}

\subsection{Multiview unsupervised learning: crosslingual word embedding}

We next seek to learn word representations that reflect word similarity,
and the multiview approach trains on (English, German) word pairs, 
hoping that information is transferred in the latent subspace.

\phantom{a}
\vspace{-2.5em}

\paragraph{Dataset.}
We obtained 36K pairs of English-German word as training examples from the parallel news commentary corpora \citep[WMT 2012-2018,][]{NewsCorpus}, 
using the word alignment method from \citet{DyeChaSmi13} and \citet{FastAlign}.
Based on the corpora we also built a bilingual dictionary, 
where each English word is matched with the (unique) German word that has been most frequently aligned to it.
The raw word embedding ($x_i$ and $y_i$) used the pretrained monolingual 300-dimensional word vectors from fastText \citep{GraBojGupetal18,FastText}.

The evaluation was conducted on two commonly used  datasets \citep{LevRei15,WordSimData}:
\textbf{a)} multilingual WS353 contains 353 pairs of English words, and their translations to German, Italian and Russian, that have been assigned similarity ratings by humans. 
It was further split into multilingual WS-SIM and multilingual WS-REL which measure similarity and relatedness between word pairs, respectively;
\textbf{b)} multilingual SimLex999 consists of 999 English word pairs and their translations.

\vspace{-0.6em}
\paragraph{Algorithms.}

All methods used multilayer perceptrons with ReLU activation.
ProxNet used the input reconstruction error as the ultimate objective.
As a result, \underline{\emph{DCCAE is exactly the RRM variant}}.
A validation set was employed to select the hidden dimension $h$ for $f$ and $g$ from $\{0.1, 0.3, 0.5, 0.7, 0.9\} \times 300$,
the regularization parameter $\lambda$,
and the depth and layer width from 1 to 4 and $\{256, 512, 1024, 2048\}$, respectively.
We searched the mini-batch size in $\{100, 200, 300, 400\}$.
We also compared with linear CCA \citep{FarDye14}.
At test time, the (English, German) word pairs from the test set were fed to the four multiview based models,
extracting the English and German word representations.
Then the cosine similarity can be computed between all pairs of monolingual words in the test set (English and German), 
and we reported in Table~\ref{tab:spearman} the Spearman's correlation between the model's ranking and human's ranking. 

\vspace{-0.8em}
\paragraph{Results.}

Clearly, ProxNet always achieves the highest or close to highest Spearman's correlation on all test sets and for both English (EN) and German (DE).
We also included a baseline which only  uses the monolingual word vectors. 
CL-DEPEMB is from \citet{Vulic17}, 
and the paper only provided  the results for

\begin{table*}[t]
	\caption{Spearman's correlation for word similarity. 
	Following \citet{WanAroLivBil15},
    for each algorithm, 
    the model with the highest Spearman's correlation on the 649 tuning bigram pairs was selected.}
	\label{tab:spearman}
	\centering
	\begin{tabular}{r|cccccccc}
		\toprule  
		&\multicolumn{2}{c}{WS-353} &\multicolumn{2}{c}{WS-SIM} &\multicolumn{2}{c}{WS-REL} &\multicolumn{2}{c}{SimLex999} \\
		\cmidrule{2-3} \cmidrule{4-5} \cmidrule{6-7} \cmidrule{8-9}
		& EN    & DE     & EN     & DE     & EN     & DE    & EN    & DE \\ 
		\midrule
		Baseline  & 73.4 & 52.7  & 77.8  & 63.3  & 67.7  & 44.2 & 37.2 & 29.1     \\ 
		LinCCA & 73.8 & 68.5  & 76.1  & 73.0  & 67.0  & 62.9 & 37.8 & 43.3     \\ 
		DCCA      & 73.9 & 69.1  & \textbf{78.7}  & 74.1  & 66.6  & 64.7 & 38.78 & 43.29     \\ 
		DCCAE / RRM     & 72.4 & \textbf{69.7}  & 75.7  & 74.7  & 65.9  & 64.2 & 36.7 & 41.8     \\ 
    	ProxNet   & \textbf{75.4} & 69.2  & 78.3  & \textbf{75.4}  & \textbf{71.0}  & \textbf{66.8} & \textbf{40.0} & \textbf{44.2}     \\ 
		CL-DEPEMB &  -    &   -    &   -    &    -   &   -    &    -  & 35.6 & 30.6     \\
		\bottomrule
	\end{tabular}
	\vspace{-2.0em}
\end{table*}
\begin{minipage}[t]{\textwidth}
	\begin{minipage}[t]{0.58\textwidth}
		\centering
		\captionof{table}{Test accuracy for sequence classification. ``len'' stands for the median length of the sequences.}		
		\label{tab:lstm_result}
		\setlength\tabcolsep{2pt}
		\begin{tabular}{r|rr|ccc}
		\toprule
		& \#train	& len & LSTM             & AdvLSTM              & ProxLSTM  \\ 
		\midrule
		JV  & 225  & 15 & 94.02 \scriptsize$\pm$0.72 
		& 94.96 \scriptsize$\pm$0.44 
		& \textbf{95.52} \scriptsize$\pm$0.63  \\
		HAR  & 6.1k & 128 & 89.75 \scriptsize$\pm$0.59
		& \textbf{92.01} \scriptsize$\pm$0.18  
		& \textbf{92.08} \scriptsize$\pm$0.23  \\
		AD & 5.5k & 39 & 96.32 \scriptsize$\pm$0.55
		& 97.45 \scriptsize$\pm$0.38
		& \textbf{97.99} \scriptsize$\pm$0.29   \\ 
		IMDB  & 25k & 239 & 92.65 \scriptsize$\pm$0.04  
		& 93.65 \scriptsize$\pm$0.03  
		& \textbf{94.16} \scriptsize$\pm$0.11   \\
		\bottomrule
		\end{tabular}
	\end{minipage}
	\hfill
	\begin{minipage}[t]{0.4\textwidth}
		\centering
		\captionof{figure}{t-SNE embedding of HAR dataset (best viewed in color)}
		\includegraphics[width=\textwidth, height=2.4cm]{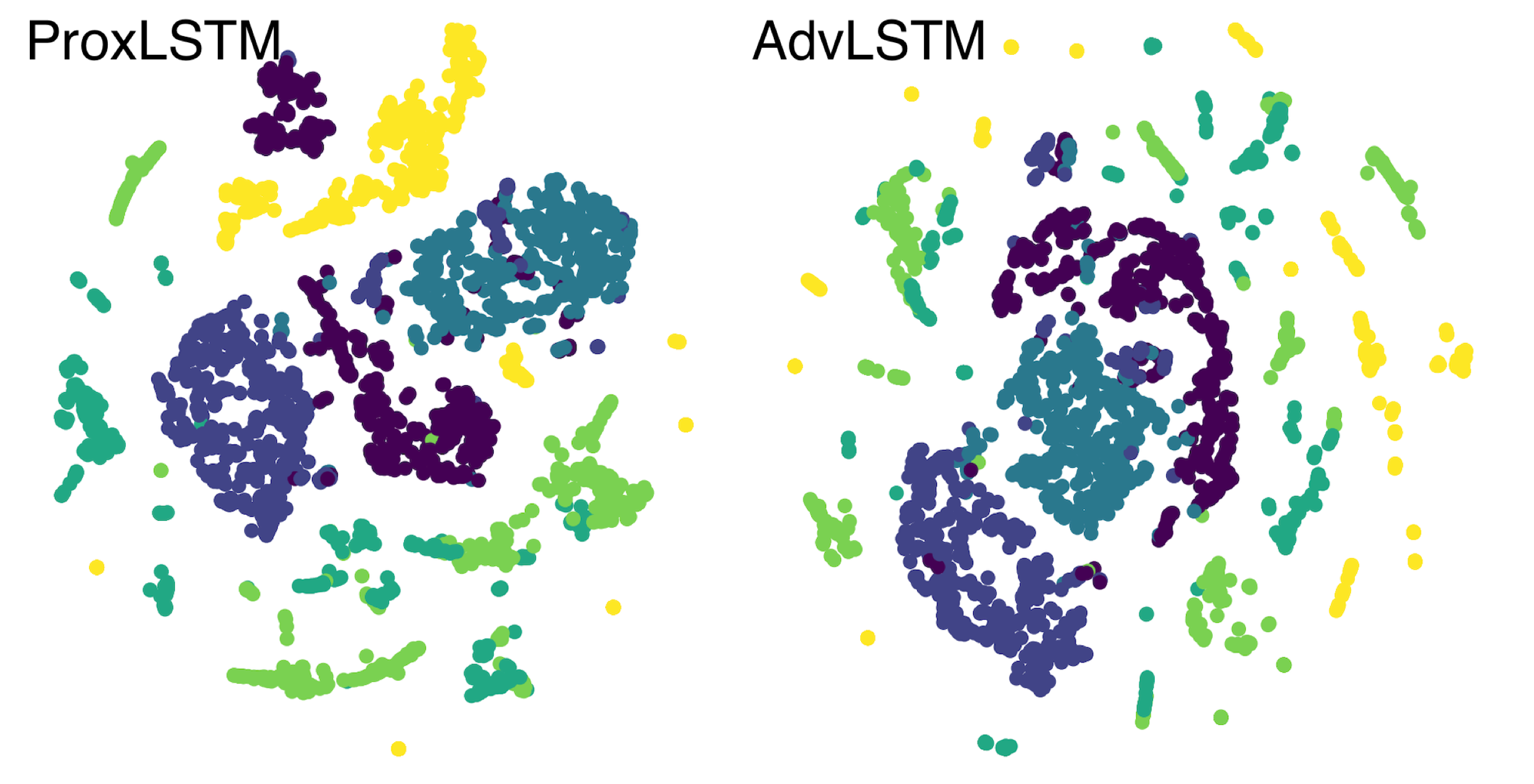}
		\label{fig:lstm_tsne}
	\end{minipage}
\end{minipage}

SimLex999 with no code made available.
It can be observed from Table~\ref{tab:spearman} that multiview based methods achieved more significant improvement over the baseline on German data than on English data.
This is not surprising, because the presence of multiple views offers an opportunity to transfer useful information from other views/languages. 
Since the performance on English is generally better than that of German,
more improvement is expected on German.
\vspace{-0.3em}
\subsection{Robust training for recurrent networks}
\label{sec:exp_adversarial}
\vspace{-0.3em}
We now present the experimental results of robust training for LSTMs as described in Section \ref{sec:lstm}. 
\vspace{-0.85em}
\paragraph{Datasets.}
We tested on four sequence datasets:
Japanese vowels \citep[JV,][]{ShiKudToy99} which contains time series data for speaker recognition based on uttered vowels;
Human Activity Recognition \citep[HAR,][]{AngGhiOne13} which classifies activity;
Arabic Digits \citep[AD,][]{HamBed10} which recognizes digits from speeches; and IMDB \citep{MaaDalPha11}, a large movie review dataset for sentiment classification.
Table \ref{tab:lstm_result} presents the training set size and \emph{median} sequence length.


\vspace{-0.85em}
\paragraph{Algorithms.}

We compared ProxLSTM with two baselines: 
vanilla LSTM and the adversarial training of LSTM \citep{MiyDaiGoo17},
which we will refer to as AdvLSTM.
For JV, HAR, AD datasets, the base models are preceded by a CNN layer, 
and succeeded by a fully connected layer. 
The CNN layer consists of kernels sized 3, 8, 3 and contains 32, 64, 64 filters for the JV, HAR, AD datasets, respectively. 
LSTM used 64, 128, 64 hidden units for these three datasets, respectively. 
All these parameters were tuned to optimize the performance of vanilla LSTM,
and then shared with ProxLSTM and AdvLSTM for a fair comparison.
We first trained the vanilla LSTM to convergence, 
and used the resulting model to initialize AdvLSTM and ProxLSTM. 
For IMDB, we first trained AdvLSTM by following the settings in \citet{MiyDaiGoo17}, 
and then used the result to initialize the weights of ProxLSTM.
All settings were evaluated 10 times to report mean and standard deviation. 
\vspace{-0.85em}
\paragraph{Results.}
From Table \ref{tab:lstm_result},
it is clear that adversarial training improves test accuracy,
and ProxLSTM promotes the performance even more than AdvLSTM. 
Since the accuracy gap is lowest on the HAR dataset,
we also plotted the t-SNE embedding of the features from the \emph{last time step} for HAR. 
As Figure \ref{fig:lstm_tsne} shows,
the representation learned by ProxLSTM is better clustered than that of AdvLSTM, especially the yellow class. 
This further indicates that ProxLSTM learns better latent representations than AdvLSTM by applying proximal mapping.
Plots for other datasets are in \S \ref{sec:app_prox_lstm}.


\vspace{-0.85em}
\paragraph{Conclusion.}
In this paper, we proposed using proximal mapping as a new primitive in deep networks to explicitly encode the prior for end-to-end training. 
Connection to existing constructs in deep learning are shown.
The new model is extended to multiview learning and robust RNNs, and its effectiveness is demonstrated in experiments.
Future work will apply it to reinforcement learning with knowledge transfer.

{
\bibliography{paper.bbl}
\bibliographystyle{xinhua_num}
\clearpage
}

\appendix
\begin{center}
\huge\bf Supplementary Material
\end{center}

\vskip 1em
\noindent
{\color{blue}
\textbf{
All code and data are available anonymously, with no tracing, at
\begin{center}
    \url{https://github.com/learndeep2019/ProxNet}.
\end{center}
}}

\section{Relationship with OptNet and Implicit Differentiation Based Learning}
\label{sec:compare_optnet}

Given a prediction model such as linear model, energy-based model, kernel function, deep neural network, etc,
a loss function is needed to measure the quality of its prediction against the given ground truth.
Although surrogate losses had been popular in making the loss convex,
recently it is often observed that directly comparing the prediction of the model, 
typically computed through an argmin optimization (or argmax),
against the ground truth under the true loss of interest can be much more effective.
The error signal is originated from the \emph{last step} through the argmin,
and then backpropagated through the model itself for training.
For example, Amos et al used it to train input convex neural networks at ICML 2017,
\citet{BelYanMac17} used it to train a structured prediction energy network,
and \citet{BraStrSch13} used it to train an energy-based model for time-series imputation.
Other works include \citet{StoRopEis11,GooMirCouetal13}, etc.
A number of implicit and auto-differentiation algorithms have been proposed for it,
\eg, \citet{Domke10,Domke12,BayPeaRadetal17,AmoKol17,GouFerCheetal16}.

Other uses of such differentiable optimization have been found in 
learning attention models \citep{NicBlo17},
meta-learning to differentiate through the base learning algorithm \citep{BerHenToretal19,LeeMajRavetal19},
or to train the generator in a generative adversarial model by optimizing out the discriminator \citep{MetPooPfaetal17},
or for end-to-end planning and control \citep{AmoRodSacetal18}.
In all these cases, differentiable optimization is used as an algorithm to train a \emph{given} component within a multi-component learning paradigm.
But each component itself has its own pre-fixed model and parameterization.

To the best of our knowledge,
OptNet \citep{AmoKol17} proposed for the first time using optimization as a \emph{layer} of the deep neural network,
hence extending the model itself.
However, it focused on efficient algorithms for differentiation\footnote{Although the original paper only detailed on quadratic optimization mainly for the efficient GPU implementation, it is conceptually applicable to general nonlinear optimization.
Such extensions have been achieved in \citet{Agrawaletal19}.}
and the general framework of optimization layer was demonstrated by using standard operations such as total variation denoising,
which bears resemblance to task-driven dictionary learning \citep{BagBra08,MaiBacPon12}.
It remains unclear how to leverage the general framework of OptNet to flexibly model a broad range of structures,
while reusing the existing primitives in deep learning (like our extension of LSTM in Section \ref{sec:lstm}).

This is achieved by ProxNet.
Although ProxNet also inserts a new layer, it provides \emph{concrete and novel} ways to model structured priors in data through proximal mapping.
Most aforementioned works use differentiable optimization as a learning algorithm for a \emph{given} model,
while ProxNet uses it as a first-class modeling construct within a deep network.
Designing the potential function $f$ in \eqref{eq:def:prox_map} can be highly nontrivial,
as we have demonstrated in the examples of dropout, kernel warping, multiview learning, and LSTM.

\citet{Rajeswaranetal19} used proximal mapping for the inner-level optimization of meta-learning, which constitutes a bi-level optimization.
Their focus is to streamline the optimization using implicit gradient,
while our goal, in contrast, is to use proximal mapping to learn structured data representations.

We note that despite the similarity in titles, 
\citet{WanFidUrt16} differs from our work as it applies proximal mapping in a solver to perform inference in a graphical model,
whose cliques are neural networks.
The optimization process \emph{happens to} be analogous to a recurrent net, 
interlaced with proximal maps,
and similar analogy has been drawn between the ISTA optimization algorithm and LSTM \citep{ZhoDiDuetal18}.
We instead use proximal map as a first-class construct/layer in a deep network.

\section{Connecting ProxNet with Meta-learning}
\label{sec:app_meta_learning}

In view that ProxNet is a bi-level optimization and the $z$ in \eqref{eq:def:prox_map} may consist of the embeddings of input objects in \emph{mini-batches},
we can interpret ProxNet from a meta-learning perspective.
In particular,
each mini-batch corresponds to a ``task'' (or dataset, episode, etc) in the standard meta-learning terminology,
and the regularization term corresponds to the task-specific base learner inside each episode of the meta learner.
Naturally, the preceding layers serve as the meta-parameters subject to meta-learning.
For example, \cite{KocZemSal15,VinBluLilKavetal16,SneSweZem17} used simple metric-based nearest neighbor,
\cite{FinLev18,FinAbbLev17} optimized standard learning algorithms iteratively,
and \cite{BerHenToretal19,LeeMajRavetal19} leveraged closed-form solutions for base learners.
Explicit learning of learner's update rule was investigated in \cite{HocYouCon01,Andrychowiczetal16,RavLar17}.
In this sense, ProxNet extends meta-learning to \emph{unsupervised} base learners.

We emphasize that ProxNet only leverages the idea and technique in meta-learning.
It is beyond our paper to address existing challenges in meta-learning itself.

\paragraph{Detailed description}
The conventional meta-learning has a meta-parameter $p$, 
and each base-learner (for each task) has its own base-parameters $w$.  
Then by Equation (1) of the paper 

Aravind Rajeswaran, Chelsea Finn, Sham Kakade, Sergey Levine. 
\textit{Meta-Learning with Implicit Gradients}.  
Neural Information Processing Systems (NeurIPS), 2019, 

the bi-level optimization in meta-learning can be set up as (``perf'' for ``performance''):
\begin{align}
\label{eq:obj_meta_learning}
    \min_p \sum_i  \text{Test-perf} \rbr{ \argmin_w \text{Training-perf} (w, p, \Dcal_i^{train}), p, \Dcal_i^{test}}.
\end{align}

Here $\Dcal_i^{train}$ and $\Dcal_i^{test}$ are the training and test data for task $i$, respectively.  
Now we can establish the one-to-one correspondence between \eqref{eq:obj_meta_learning} and ProxNet in the context of multiview learning.
Please refer to Section \ref{sec:multiview} for notations, 
especially Equations \eqref{eq:obj_cca} and \eqref{eq:prox_multiview}.
\begin{itemize}
    \item 
    $p$: the union of i) the feature extractors $f$ and $g$ for the two views, and ii) the downstream supervised layers.  Only the former ($f$ and $g$) is used in the inner training ($\argmin_w$),
    which transforms the raw data into the input of the proximal layer.
    \item 
    $w$: the $U$ and $V$ projection directions used by CCA;
    \item 
    $\Dcal_i^{train}$: the $i$-th mini-batch $\{(x_i,y_i)\}_{i=1}^n$;
    \item 
    Training-perf$(w, p, \Dcal_i^{train}) = \min\limits_{P,Q} \frac{\lambda}{2n} \nbr{P - X}_F^2 + \frac{\lambda}{2n} \nbr{Q -Y}_F^2 - \tr(U^\top PQ^\top V)$,
    where $X = (f(x_1),\ldots, f(x_n))$ and $Y = (g(y_1),\ldots, g(y_n))$.
    That is, for any given projection directions $U$ and $V$ (\ie, $w$),
    what is the minimal denoising objective,
    which combines the displacement (Frobenius norm) and the CCA objective (correlation between the projections);
    \item 
    $\Dcal_i^{test}$: the $i$-th mini-batch (same as $\Dcal_i^{train}$);
    \item
    Test-perf: pass $\Dcal_i^{test}$ through $f$ and $g$, followed by denoising based on the trained $w = (U, V)$:
    $\mathop{{\color{blue}\argmin}}\limits_{P,Q} \frac{\lambda}{2n} \nbr{P - X}_F^2 + \frac{\lambda}{2n} \nbr{Q -Y}_F^2 - \tr(U^\top PQ^\top V)$,
    and finally apply the supervised layers to measure the test performance.
\end{itemize}

So ProxNet effectively corresponds to a base-learner of multiview denoising. 
It extends the common meta-learning practice in two ways: 
\begin{itemize}
    \item the base-learner is unsupervised;
    \item the training and test performance employ different tasks (denoising versus error).
\end{itemize}  
The latter is quite a valid learning paradigm: 
the training phase extracts useful representations as parameterized by $U$ and $V$, 
and then the product ($U$ and $V$) is evaluated on the test data by computing their projections, followed by a supervised loss.  
Since mini-batch sizes are very small (also intended to keep the optimization efficient), 
it can be considered as a few-shot learning.  
Surely the algorithm does not have to be restricted to mini-batches that are drawn iid; 
different mini-batches can employ bona-fide different learning tasks.

\section{Connecting Proximal Mapping to Kernel Warping}
\label{sec:app_warping}

The graph Laplacian on a function $f$ is
$\sum_{ij} w_{ij} (f(x_i)  -  f(x_j))^2$,
where $f(x_i)  -  f(x_j)$ is bounded and linear in $f$.
Parameterizing an image as $I(\alpha)$ where $\alpha$ is the degree of rotation/translation/etc,
transformation invariance favors a small magnitude of $\frac{\partial}{\partial \alpha} |_{\alpha=0} f(I(\alpha))$, again a bounded linear functional.
By Riesz representation theorem,
a bounded linear functional can be written as $\inner{z_i}{f}_\Hcal$ for some $z_i  \in  \Hcal$.
We will refer to $z_i$ as an invariance representer,
and suppose we have $m$ such invariances.

In order to respect the desired invariances,
\citet{MaGanZha19} proposed a warped RKHS $\Hcal^\circ$ consisting of the same functions in the original $\Hcal$,
but redefining the norm and the corresponding kernel by
\begin{align}
\nbr{f}_{\Hcal^\circ}^2 :=& \nbr{f}_\Hcal^2 + \sum\nolimits_{i=1}^m \inner{z_i}{f}^2_\Hcal
\end{align}

This leads to a new RKHS consisting of the same set of functions as $\Hcal$,
but its inner product warped into
\begin{align}
    \inner{f}{g}_{\Hcal^\circ} := \inner{f}{g}_{\Hcal} + \sum\nolimits_{i=1}^m \inner{f}{z_i}_\Hcal \inner{g}{z_i}_\Hcal,
\end{align}
and its kernel is warped into
\begin{align}
    k^\circ(x_1, x_2) =& k(x_1,x_2) - z(x_1)^\top K_Z z(x_2),
\end{align}
where $z(x) = (z_1(x), \ldots, z_m(x))^\top$.
Then replacing $k(x,\cdot)$ by $k^\circ(x, \cdot)$ results in a new invariant representation.
Such a warping can be applied to all layers in,
\eg, deep convolutional kernel networks \citep[CKNs,][]{BieMai19},
instilling invariance with respect to preceding layer's output.

The major limitation of this method, however, is that the invariances have to be modeled by the square of a linear form --- $\inner{z_i}{f}^2_\Hcal$ --- in order to make $\nbr{f}_\Hcal^2 + \sum\nolimits_{i=1}^m \inner{z_i}{f}^2_\Hcal$ a norm square,
precluding many interesting invariances such as total variation $f \mapsto \int |f'(x)| \rmd x$.

Interestingly, this can be achieved by simply reformulating kernel warping as proximal mapping.
To this end, recall that a Euclidean embedding maps $f \in \Hcal$ to a real vector $\ftil$, 
such that 
$\langle \ftil, \htil \rangle \approx \inner{f}{h}_\Hcal$
for all $f, h \in \Hcal$.
A commonly used formula for embedding is the Nystr\"{o}m approximation \citep{WilSee00b}.
Using $p$ samples $W := \{\omega_i\}_{i=1}^p$ drawn \iid\ from $\Xcal$,
we derive an embedding of $f \in \Hcal$ as follows, 
ensuring that $\langle\ftil,\htil\rangle \approx \inner{f}{h}_\Hcal$ for all $f, h \in \Hcal$:

\vspace{-1.7em}
\begin{align*}
	\ftil := K_W^{-1/2} f_W, \where K_W := (k(\omega_i,\omega_j))_{ij} \in \RR^{p \times p}, \quad f_W := (f(\omega_1), \ldots, f(\omega_p))^\top \in \RR^p.
\end{align*}

Let $\tilde{\varphi}(x)$ be the embedding of $k(x, \cdot)$,
and $\Ztil := (\ztil_1, \ldots, \ztil_m)$ where $\ztil_i$ is the embedding of the invariance representer $z_i$.
Then \citet{MaGanZha19} showed that the Euclidean embedding of $k^\circ(x,\cdot)$ can be written as
\begin{align}
\label{eq:fa_warping}
	(I + \Ztil \Ztil^\top)^{-1/2} \tilde{\varphi}(x).
\end{align}
Now to apply proximal map,
it is natural to set $L(f) = \frac{1}{2} \sum_{i=1}^m \inner{z_i}{f}^2_\Hcal$ to enforce invariance.
Then the proximal map $\Psf_L(k(x,\cdot))$ for the representer $k(x,\cdot)$ with $\lambda = 1$ is
\begin{align}
\label{eq:prox_rkhs_app}
    \Psf_L(k(x,\cdot))
    &= 
    \argmin_{f \in \Hcal} \cbr{L(f) + \smallfrac{1}{2}\nbr{f - k(x,\cdot)}^2_\Hcal} 
    \\
    &= \argmin_{f \in \Hcal} \cbr{\smallfrac{1}{2} \sum\nolimits_{i=1}^m \inner{z_i}{f}^2_\Hcal + \smallfrac{1}{2}\nbr{f - k(x,\cdot)}^2_\Hcal} \\
    &= (I + Z Z^\top)^{-1} k(x,\cdot).
\end{align}
Its Euclidean embedding can be obtained by replacing $z_i$ with $\ztil_i$,
and $k(x,\cdot)$ with $\tilde{\varphi}(x)$:
\begin{align}
\label{eq:prox_rkhs_eucl}
    \argmin_{v \in \RR^p} \cbr{\smallfrac{1}{2} \sum\nolimits_{i=1}^m \inner{\ztil_i}{v}^2 + \smallfrac{1}{2}\nbr{v - \tilde{\varphi}(x)}^2} 
    = 
    (I + \Ztil \Ztil^\top)^{-1} \tilde{\varphi}(x).
\end{align}
This is almost the same as that from kernel warping in \eqref{eq:fa_warping},
except for the exponent on $I + \Ztil \Ztil^\top$.
In practice, we observed that it led to little difference,
and the result of proximal mapping using Gaussian kernel and flat-gradient invariance is shown in Figure \ref{fig:twomoon}.
That is, $L(f) = \frac{1}{2}\sum_i \nbr{\grad f(x_i)}^2$.
Trivially, CKNs can now leverage nonlinear invariances such as total variation by using a nonlinear  regularizer $L$ in \eqref{eq:prox_rkhs_app}.

%


\section{Simulations for Connecting Proximal Mapping to Dropout}
\label{sec:app_dropout}

We now use the two-moon dataset to verify that only small differences arise if dropout is implemented by proximal mapping in Section \ref{sec:shallow}, as opposed to the adaptive regularization in \eqref{eq:reg_dropout}.
Suppose the $i$-th training examples is $x_i \in \RR^d$ with label $y_i \in \{-1,1\}$.
The $j$-th feature of $x_i$ is denoted as $x_{ij}$.
Employing logistic loss,
the adaptive regularization view of dropout by \citet{WagWanLia13} can be written as
\begin{align}
	\beta_* &:= \min_{\beta \in \RR^d} \cbr{\frac{1}{n} \sum_{i=1}^n \log (1+\exp(-y_i \beta^\top x_i)) + \mu \sum_j a_j \beta_j^2},
\end{align}
where $a_j = \frac{1}{n} \sum_{i=1}^n p_i (1 - p_i) x_{ij}^2,
	\quad p_i = (1 + \exp(-\beta^\top x_i))^{-1}$.

Our proximal map is defined as
\begin{align}
	\Psf_R(x) &= \argmin_{z \in \RR^d} \cbr{\frac{\lambda}{2} \nbr{z - x}_2^2 + \sum_j b_j z_j^2},
\end{align}
where $b_j = \frac{1}{n} \sum_{i=1}^n q_i (1 - q_i) x_{ij}^2,
	\quad q_i = (1 + \exp(-z^\top x_i))^{-1}$.

And the output layer is trained by
\begin{align}
	\alpha_* &:= \min_{\alpha \in \RR^d} \cbr{\frac{1}{n} \sum_{i=1}^n \log (1+\exp(-y_i \alpha^\top \Psf_R(x_i)) + c \nbr{\alpha}^2}.
\end{align}

\begin{figure*}[htbp!]
    \centering
    \subfloat[$\lambda = 0.5$, $\mu = 0.1$, and $c = 0.2 \lambda^2 \mu$]
    {\includegraphics[width=0.45\textwidth]{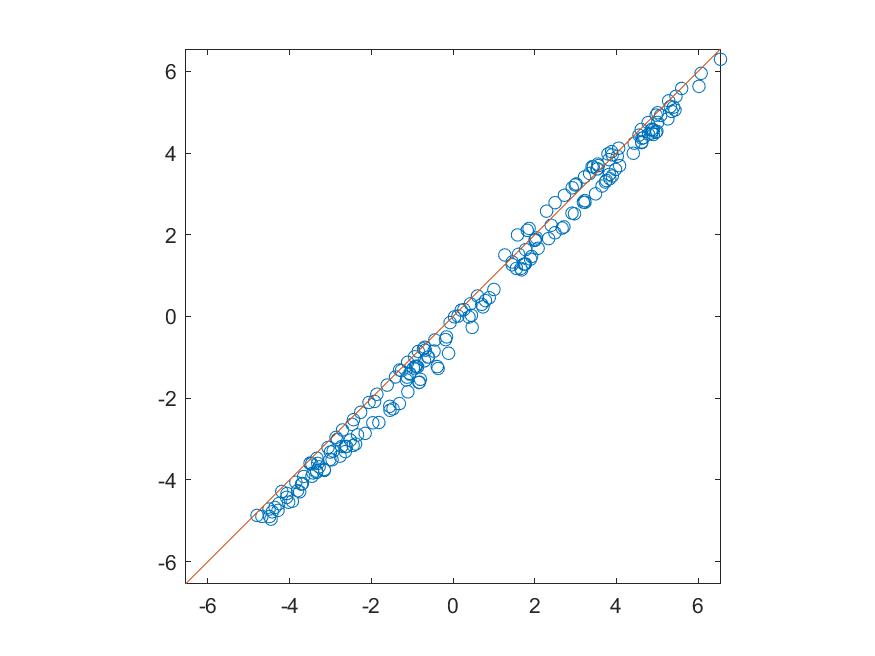}
    \label{fig:dropout_1}
    }
    ~ 
    \centering
    \subfloat[$\mu = 0.1$, $\lambda = 0.1$, and $c = 15 \lambda^2 \mu$]
    {\includegraphics[width=0.45\textwidth]{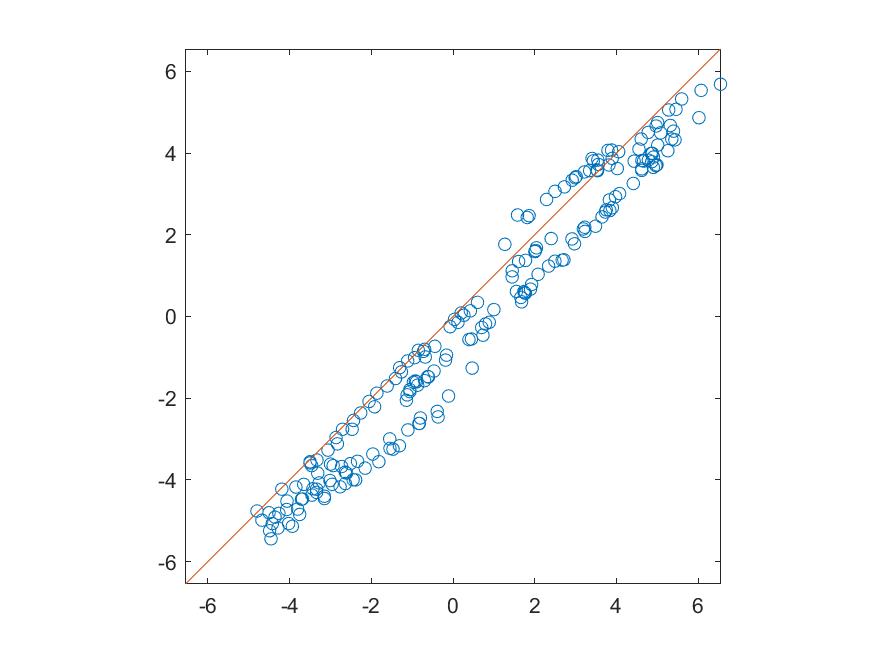}
    \label{fig:dropout_2}
    }
    \caption{Scatter plot of $\alpha_*^\top P_R(x_i)$ ($y$-axis for proximal mapping) versus $\beta_*^\top x_i$ ($x$-axis for dropout)}
    \label{fig:dropout}
\end{figure*}

To demonstrate that the two methods yield similar discriminant values,
we produce a scatter plot of $\alpha_*^\top P_R(x_i)$ (for proximal mapping) versus $\beta_*^\top x_i$ (for dropout).
Figure \ref{fig:dropout} shows the result for two example settings.
Clearly, the two methods produce similar discriminant values for all training examples.
The Matlab code is also available on GitHub.

\section{ProxNet for Multiview Learning}
\label{sec:app_multiview}

Most multiview learning algorithms are based on CCA,
which most commonly involves only two views.
It is in fact not hard to extend it to more than two views.
For example, \citet{Horst19} proposed that given $J$ centered views $X_j \in \RR^{N \times d_j}$ for $j \in [J]$, where $N$ is the number of training examples and $d_j$ is the dimensionality of the $j$-th view,
the generalized CCA (GCCA) can be written as the following optimization problem
\begin{align}
\label{eq:obj_CCA_general}
L(\{X_j\}_{j=1}^J) := \min \sum_{j=1}^J \nbr{G - X_j U_j}_F^2,
\end{align}
where $G \in \RR^{N \times r}, \ U_j \in \RR^{d_j \times r}, \ G^\top G = I$.
Intuitively, it finds a linear transformation $U_j$ for each view,
so that all views can be transformed to a similar core $G$.
Furthermore, $G$ needs to be orthonormal, to avoid mode collapse.
The optimal value, denoted as $L(\{X_j\})$,
will be used as the $L$ function in \eqref{eq:prox_multiview}.

Furthermore, given $\{X_j\}$, \eqref{eq:obj_CCA_general} can be optimized efficiently in closed form based on generalized eigenvalues \citep{RasVanAro15,Horst19,Bentonetal19}.
Based on the optimal solution of $G$ and $\{U_j\}$, 
the derivative of $L(\{X_j\})$ in $\{X_j\}$ can be directly computed by Danskin's theorem.

\section{Backpropagation Through Time for Adversarial LSTM}
\label{sec:bp_lstm}

To concentrate on backpropagation,
we assume that the ultimate objective $J$ only depends only on the output of the last time step $T$, \ie, $h_T$.
Extension can be easily made to the case where each step also contributes to the overall loss.
From the final layer, we get $\frac{\partial J}{\partial h_T}$.
Then we can get $\frac{\partial J}{\partial h_{T-1}}$ and 
$\frac{\partial J}{\partial c_{T-1}}$ as in the standard LSTM ($G_T$ in the final layer can be ignored and 
$\frac{\partial J}{\partial c_T} = 0$). 
In order to compute the derivatives with respect to the weights $W$ in the LSTMs, we need to recursively compute 
$\frac{\partial J}{\partial h_{t-1}}$ and 
$\frac{\partial J}{\partial c_{t-1}}$,
given $\frac{\partial J}{\partial h_{t}}$ and 
$\frac{\partial J}{\partial c_{t}}$.
Once they are available, then
\begin{align}
\frac{\partial J}{\partial W} 
= \sum_{t=1}^T \cbr{\underbrace{\frac{\partial J}{\partial h_t}}_{\text{by } \eqref{eq:rec_ell_y}} 
	\underbrace{\frac{\partial}{\partial W} h_t(c_{t-1}, h_{t-1}, x_t)}_{\text{standard LSTM}} + \underbrace{\frac{\partial J}{\partial c_t}}_{\text{by } \eqref{eq:rec_ell_c}} \underbrace{\frac{\partial}{\partial W} c_t(c_{t-1}, h_{t-1}, x_t)}_{\text{standard LSTM}}},
\end{align}
where the two $\frac{\partial}{\partial W}$ on the right-hand side are identical to the standard operations in LSTMs.
Here we use the Jacobian matrix arrangement for partial derivatives,
\ie,
if $f$ maps from $\RR^n$ to $\RR^m$,
then $\frac{\partial f(x)}{\partial x} \in \RR^{m \times n}$. 

Given $\frac{\partial J}{\partial c_{t}}$, we can first compute 
$\frac{\partial J}{\partial s_t}$ and $\frac{\partial J}{\partial G_t}$ based on the proximal map,
and the details will be provided in Section \ref{sec:grad_prox}.
Given their values,
we now compute
$\frac{\partial J}{\partial h_{t-1}}$ and 
$\frac{\partial J}{\partial c_{t-1}}$.
Firstly,
\begin{align}
\label{eq:rec_ell_y}
\frac{\partial J}{\partial h_{t-1}} = 
\underbrace{\frac{\partial J}{\partial h_t}}_{\text{by recursion}}
\underbrace{\frac{\partial h_t}{\partial h_{t-1}}}_{\text{std LSTM}}
+ \underbrace{\frac{\partial J}{\partial G_t} \frac{\partial G_t}{\partial h_{t-1}}}_{\text{by }\eqref{eq:ell_G_y}}
+ \underbrace{\frac{\partial J}{\partial s_t}}_{\text{by } \eqref{eq:ell_s}}
\underbrace{\frac{\partial s_t}{\partial h_{t-1}}}_{\text{std LSTM}}.
\end{align}
The terms $\frac{\partial h_t}{\partial h_{t-1}}$ and $\frac{\partial s_t}{\partial h_{t-1}}$ are identical to the operations in the standard LSTM.
The only remaining term is in fact a directional second-order derivative,
where the direction $\frac{\partial J}{\partial G_t}$ can be computed from from \eqref{eq:ell_G}:
\begin{align}
\label{eq:ell_G_y}
\frac{\partial J}{\partial G_t} \frac{\partial G_t}{\partial h_{t-1}} &=
\frac{\partial J}{\partial G_t} \frac{\partial^2 }{\partial x_t \partial h_{t-1}} s_t(c_{t-1}, h_{t-1}, x_t) \\
&= \frac{\partial}{\partial h_{t-1}} \inner{\underbrace{\frac{\partial J}{\partial G_t}}_{\text{by } \eqref{eq:ell_G}}}{\frac{\partial}{\partial x_t} s_t(c_{t-1}, h_{t-1}, x_t)}.
\end{align}
Such computations are well supported in most deep learning packages, such as PyTorch.
Secondly, 
\begin{align}
\label{eq:rec_ell_c}
\frac{\partial J}{\partial c_{t-1}} = 
\underbrace{\frac{\partial J}{\partial h_t}}_{\text{by recursion}}
\underbrace{\frac{\partial h_t}{\partial c_{t-1}}}_{\text{std LSTM}}
+ \underbrace{\frac{\partial J}{\partial G_t} \frac{\partial G_t}{\partial c_{t-1}}}_{\text{by }\eqref{eq:ell_G_c}}
+ \underbrace{\frac{\partial J}{\partial s_t}}_{\text{by } \eqref{eq:ell_s}}
\underbrace{\frac{\partial s_t}{\partial c_{t-1}}}_{\text{std LSTM}}.
\end{align}
The terms $\frac{\partial h_t}{\partial c_{t-1}}$ and $\frac{\partial s_t}{\partial c_{t-1}}$ are identical to the operations in the standard LSTM.
The only remaining term is in fact a directional second-order derivative:
\begin{align}
\label{eq:ell_G_c}
\frac{\partial J}{\partial G_t} \frac{\partial G_t}{\partial c_{t-1}} &=
\frac{\partial J}{\partial G_t} \frac{\partial^2 }{\partial x_t \partial c_{t-1}} s_t(c_{t-1}, h_{t-1}, x_t)\\
&= \frac{\partial}{\partial c_{t-1}} \inner{\underbrace{\frac{\partial J}{\partial G_t}}_{\text{by } \eqref{eq:ell_G}}}{\frac{\partial}{\partial x_t} s_t(c_{t-1}, h_{t-1}, x_t)}.
\end{align}

\subsection{Gradient Derivation for the Proximal Map}
\label{sec:grad_prox}

We now compute the derivatives involved in the proximal operator,
namely $\frac{\partial J}{\partial s_t}$ and $\frac{\partial J}{\partial G_t}$.
For clarify, let us omit the step index $t$, 
set $\delta = \sqrt{\lambda}$ without loss of generality, 
and denote
\begin{align}
J = f(c), \where c := c(G,s) := (I + G G^\top)^{-1} s.
\end{align}
We first compute $\partial J / \partial s$ which is easier.
\begin{align}
\Delta J :=& f(c(G, s+\Delta s)) - f(c(G,s)) \\
=& \grad f(c)^\top (c(G,s+\Delta s) - c(G,s)) + o(\nbr{\Delta s})\\
=& \grad f(c)^\top (I+GG^\top)^{-1} \Delta s + o(\nbr{\Delta s}).
\end{align}
Therefore, 
\begin{align}
\label{eq:ell_s}
\frac{\partial J}{\partial s} = \grad f(c)^\top (I+GG^\top)^{-1}.
\end{align}
We now move on to $\partial J / \partial G$.
Notice
\begin{align}
\Delta J :=& f(c(G+\Delta G,s)) - f(c(G,s)) \\
=& \grad f(c)^\top (c(G+\Delta G,s) - c(G,s)) + o(\nbr{\Delta G}).
\end{align}
Since
\begin{align}
&c(G+\Delta G,s) = (I+(G+\Delta G)(G + \Delta G)^\top)^{-1} s \\
= &\sbr{(I+GG^\top)^{\half} \rbr{I + (I+GG^\top)^{-\half} (\Delta G G^\top + G \Delta G^\top) (I+GG^\top)^{-\half}}(I+GG^\top)^{\half}}^{-1} s \\
= &(I+GG^\top)^{-\half} \rbr{I - (I+GG^\top)^{-\half} (\Delta G G^\top + G \Delta G^\top) (I+GG^\top)^{-\half} + o(\nbr{\Delta G})}(I+GG^\top)^{-\half} s \\
= & c(G,s) - (I+GG^\top)^{-1} (\Delta G G^\top + G \Delta G^\top) (I+GG^\top)^{-1} s + o(\nbr{\Delta G}),
\end{align}
we can finally obtain
\begin{align}
\Delta J
&= -\grad f(c)^\top (I+GG^\top)^{-1} (\Delta G G^\top + G \Delta G^\top) (I+GG^\top)^{-1} s + o(\nbr{\Delta G}) \\
&= -\tr \rbr{\Delta G^\top (I+GG^\top)^{-1} \rbr{\grad f(c) s^\top + s \grad f(c)^\top} (I+GG^\top)^{-1} G} + o(\nbr{\Delta G}).
\end{align}
So in conclusion,
\begin{align}
\frac{\partial J}{\partial G} &= - (I+GG^\top)^{-1} \rbr{\grad f(c) s^\top + s \grad f(c)^\top} (I+GG^\top)^{-1} G \\
\label{eq:ell_G}	
&= -(a c^\top + c a^\top)G,
\where a = (I+GG^\top)^{-1}\grad f(c).
\end{align}

\section{Detailed Experimental Result}
\label{sec:app_exp}


\noindent
{\color{blue}
\textbf{
All code and data are available anonymously, with no tracing, at
\begin{center}
    \url{https://github.com/learndeep2019/ProxNet}.
\end{center}
}}

We will demonstrate the effectiveness of ProxNet on several multi-view learning tasks including  image classification, speech recognition, and crosslingual word embedding. 
Four baseline methods were selected for comparison in multi-view learning:
\begin{itemize}
    \item \textbf{Vanilla} model: a network is trained for each view without CCA regularization, and the output of the two views were combined by averaging their logits for supervised tasks. The network is trained in an end-to-end manner.
    %
    \item \textbf{DCCA} \citep{AndAroBiletal13}: a network is trained to learn a pair of highly-correlated representations for the two views, which are then used for training the subsequent supervised task. The whole model is trained in a disjoint manner. 
    \item \textbf{DCCAE} \citep{WanAroLivetal15}: same as DCCA, except that it trains an extra decoder to enforce that the learned representations can well reconstruct the input.
    \item \textbf{RRM}: connect the code/output of DCCA with a supervised classifier,
    and train it with the encoder in an end-to-end fashion.
    It also resembles ProxNet, 
    except that the regularizer $L(X, Y)$ is moved from the proximal layer to the overall objective as in \eqref{eq:obj_RRM} (\ie, no more proximal mapping).
\end{itemize}

\subsection{Multiview Supervised Learning: image recognition with sketch and photo}
\label{sec:app_sketchy}

\paragraph{Dataset.}

We first evaluated ProxNet on a large scale sketch-photo paired database -- Sketchy which consists of 12,500 photos and 75,471 hand-drawn sketches of objects from 125 classes. 
Each sample from sketch and photo is $256\times 256$ colored natural images.
To demonstrate the robustness of our method, we varied the number of classes over $\{20, 50, 100, 125\}$ by sampling a subset of classes from the original dataset.
For each class, there are 100 sketch-photo pairs.
We randomly selected 80 pairs of photo and sketch from each same class to form the training set, and then used the remaining 20 pairs for testing.

\paragraph{Implementation detail.}
Our implementation was based on PyTorch and all training was conducted on one NVIDIA GeForce 2080Ti GPU.

For all methods, we used ResNet-18 as the feature extractor.
In ProxNet, the feature extractor immediately followed by a proximal layer which has input and output dimension $d=20$.
Then a classifier which has three fully-connected layer each having 512 units was trained on the outputs of proximal layer.
The final output layer has multiple softmax units that each corresponds to the output classes.
At training time, we employed an adaptive trade-off parameter $\lambda_t = (1 + kt)\alpha_0$, where $k=0.5$ and $\alpha_0=0.1$.
RRM used the same architecture as ProxNet,
except that, instead of using the proximal layer, 
RRM moves the CCA objective (\ie, the regularizer $L(X, Y)$) to the overall objective to promote the correlation between the two views' hidden representation.

Since the Vanilla model does not promote correlation between views,
it can be adapted from RRM model by removing the regularizer from the overall objective.
\citet{AndAroBiletal13,WanAroLivetal15} trained DCCA and DCCAE in two separate steps instead of end-to-end.  
The first step learned an encoder (and decoder for DCCAE) to optimize the CCA objective, 
and the second step trained a supervised classifier based on the code. 
In our experiment, their encoders employed the same architecture as the feature extractors of ProxNet and other baselines, i.e., ResNet-18. 
For DCCAE, we built a CNN-based decoder to reconstruct the inputs.

For all methods, the loss was evaluated on the averaged logits at training time in order to be consistent with how predictions were made at test time.

We used the Ray Tune library to select the hyper-parameters for all methods,
and the selected parameters are summarized here:
\begin{table}[h]
\setlength\tabcolsep{3pt}
	\caption{Hyper-parameters for all methods on the Sketchy dataset}
	\label{tab:hyper_sketchy}
	\centering
    \begin{tabular}{@{}rccccccccc@{}}
    	\toprule
    	Hyper-parameters &  Vanilla & DCCA &  DCCAE &  RRM & ProxNet\\
    	\midrule
    	Dimension $d$    & 15 & 22 & 19 & 20 & 20\\
    	Optimizer & Adam & Adam & Adam & Adam & Adam \\
    	Learning rate  & 0.0012 & 0.0010 & 0.0009 & 0.0011 & 0.0011\\
    	Weight decay   &$10^{-4}$ &$10^{-4}$ &$10^{-4}$ &$10^{-4}$ &$10^{-4}$\\
    	\bottomrule
    \end{tabular}
\end{table}

The accuracy of all methods saturates after the mini-batch size goes above 100.
So we just used 100 for all methods to keep training efficient.

\subsection{Audio-Visual Speech Recognition}


\paragraph{Dataset.}
In this task, we aim to use learned features for speaker-independent phonetic recognition.
We experimented on the Wisconsin X-ray Micro-Beam Database (XRMB) corpus which consists of simultaneously recorded speech and articulatory measurements from 47 American English speakers and 2357 utterances. 
The two raw-input views are acoustic features (39D mel frequency cepstral coefficients (MFCCs) and their first and sencond derivatives) and articulatory features (16D horizontal/vertical displacement of 8 pellets attached to several parts of the vocal tract).
Along with the multi-view data there are phonetic labels available for classification.
To simulate the real-life scenarios and improve the model's robustness to noise, the acoustic features of a given speaker are corrupted by mixing with $\{0.2, 0.5, 0.8\}$ level of another random picked speaker's acoustic features. 
The XRMB speakers were partitioned into disjoint sets of 35/12 speakers for training and testing respectively.


\paragraph{Implementation detail.}
In \citet{WanAroLivetal15}, to incorporate contexts information, the inputs are concatenated over a $W$-frame window centered at each frame, giving $39\times W$ and $16\times W$ feature dimensions for each of the two views respectively. 
Although this delicately construed inputs freed the encoder/feature extractor from considering the time dependency within frames, we prefer a refined modeling of the sequential structure. 
Therefore, instead of concatenating features for each $W$-frame window followed by a fully connected network as in \citet{WanAroLivBil15}, 
we implementated, for all algorithms under consideration, 
a 2-layer LSTM with hidden size $256$.
The output of LSTM was passed through a fully connected layer,
projecting to a $K$-dimensional subspace.
This feature extractor significantly improved the performance of all methods.

The supervised predictor was implemented by a fully connected network of 2 hidden layers each having $256$ ReLU units, 
and a linear output layer of 41 log-softmax units. 
We used Pytorch's built-in function Connectionist Temporal Classification (CTC) loss \citep{AleSanFau06} with greedy search as the phone recognizer.
Again, all methods shared the same architecture of supervised predictor.

Both RRM and Vanilla were trained in the same way as for the Sketchy dataset in Section \ref{sec:app_sketchy}.
To train ProxNet, we employed an adaptive trade-off parameter $\lambda_t = (1 + kt)\alpha_0$, where $k=1$ and $\alpha_0=0.5$.
DCCA and DCCAE performed poorly if only the learned code/features were used for phonetic recognition. 
Therefore, we followed \cite{WanAroLivetal15} and concatenated them with the original features (39D and 16D for the acoustic and articulatory views, respectively), 
based on which a CTC-based recognizer is trained.
This improved the PER performance of DCCA and DCCAE significantly.

In the logit averaging mode, all methods were trained with a loss applied to the averaged logits.
This is the same as Section \ref{sec:app_sketchy}.
In the acoustic mode, however, a loss is applied to each view at training time based on the ground truth label.
These are both consistent with how predictions are made at test time.


%

Here we intentionally used $K$ instead of $d$ to denote the hidden dimension.
This is to avoid confusion because LSTM is used as in Figure \ref{fig:ProxNet_seq}.
For a mini-batch of size $m$ where each sequence has length $s$,
the input of the proximal layer is in fact
$m \cdot s$ examples of $K$ dimensional.
Although $m \cdot s$ may result in a large number, 
the proximal mapping can still be solved efficiently because we were able to use a larger value of $\lambda$ in this dataset. 
In addition, the computational cost for SVD on an $ms$-by-$K$ matrix is $O(msK^2)$ when $K \le ms$.
Since we used $K=20$, the quadratic dependency on $K$ did not create a computational challenge in practice.

As in the Skytch dataset, we used the Ray Tune library to select the hyper-parameters,
and the selected parameters are summarized here:
\begin{table}[h]
\setlength\tabcolsep{3pt}
	\caption{Hyper-parameters for all methods on XRMB}
	\label{tab:hyper_audio}
	\centering
    \begin{tabular}{@{}rcccccc@{}}
    	\toprule
    	Hyper-parameters &  Vanilla & DCCA &  DCCAE &  RRM & ProxNet\\
    	\midrule
    	Dimension $K$    & 12 & 20 & 20 & 18 & 20\\
    	Optimizer & Adam & Adam & Adam & Adam & Adam \\
    	Learning rate  & 0.0009 & 0.0011 & 0.0010 & 0.0013 & 0.0010\\
    	Weight decay   &0.0005 & 0.0005 & 0.0005 & 0.0005 & 0.0005\\
    	\bottomrule
    \end{tabular}
\end{table}

Line 250 made an inaccurate description of how we tuned $K$: 
``The dimension of subspace was tuned in $\{10, 20, 30, 50\}$,
and the sequence length was tuned in $\{250,500,1000\}$ for all algorithms''.
This was the setting in our preliminary experiment.
The Ray Tune library indeed allowed us to later search all parameters in a continuous space,
and so the $K$ values in Table \ref{tab:hyper_audio} can be 12 or 18.

We eventually set the sequence length to $s=1000$ for all methods,
because it consistently produced the best result,
which is not surprising because longer sequences can preserve more structure.
However, the PER saturated after the length rose beyond 1000.

Similarly, the PER of all methods leveled off after the mini-batch size grew above 32.
So we just used $m=32$ for all methods to keep training efficient.



\paragraph{Evaluation.}
For all experiments, we report the Phone error rates (PERs) which is defined as $PER = (S+D+I)/N$, where $S$ is the number of substitutions, $D$ is the number of deletions, $I$ is the number of insertions to get from the reference to the hypothesis, and $N$ is the number of phonetics in the reference. 
The PERs obtained by different methods are given in Table \ref{tab:pers}.

\subsection{Crosslingual/Multilingual Word Embedding}

In this task, we learned representation of English and German words from the paired (English, German) word embeddings for improved semantic similarity. 

\paragraph{Dataset.} 
We first built a parallel vocabulary of English and German from the parallel news commentary corpora \citep[WMT 2012-2018][]{NewsCorpus} using the word alignment method from \citet{FastAlign,DyeChaSmi13}. 
Then we selected 36K English-German word pairs, in descending order of frequency, for training.
Based on the vocabulary we also built a bilingual dictionary for testing, 
where each English word $x_i$ is matched with the (unique) German word $y_i$ that has been most frequently aligned to $x_i$. 
Unlike the setup in \citet{FarDye14} and \citet{WanAroLivetal15}, 
where word embeddings are trained via Latent Semantic Analysis (LSA) using parallel corpora, we used the pretrained monolingual 300-dimensional word embedding from  \citet{FastText} and \citet{GraBojGupetal18} as the raw word embeddings ($x_i$ and $y_i$). 

To evaluate the quality of learned word representation, 
we experimented on two different benchmarks that have been widely used to measure word similarity \citep{WordSimData,LevRei15}.
Multilingual WS353 contains 353 pairs of English words, and their translations to German, Italian and Russian, that have been assigned similarity ratings by humans.
It was further split into Multilingual WS-SIM and Multilingual WS-REL which measure the similarity and relatedness between word pairs respectively.
Multilingual SimLex999 is a similarity-focused dataset consisting of 666 noun pairs, 222 verb pairs, 111 adjective pairs, 
and their translations from English to German, Italian and Russian.

\paragraph{Baselines.}
We compared our method with the monolingual word embedding (baseline method) from fastText to show that ProxNet learned a good word representation through the proximal layer.
Since our method is mainly based on CCA, 
we also chose three competitive CCA-based models for comparison, including: 
\begin{itemize}
	\item 
		linearCCA \citep{FarDye14}, which applied a linear projection on the two languages' word embedding and then projected them into a common vector space such that aligned word pairs should be maximally correlated. 
	\item 
		DCCA \citep{LuWanBan15}, which, instead of learning linear transformations with CCA,  learned nonlinear transformations of two languages' embedding that are highly correlated.
	\item 
		DCCAE \citep{WanAroLivetal15}, which noted that there is useful information in the original inputs that is not correlated across views. Therefore, they not only projected the original embedding into subspace, but also reconstructed the inputs from the latent representation. 
	\item 
		CL-DEPEMB \citep{Vulic17}, a novel cross-lingual word representation model which injects syntactic information through dependency-based contexts into a shared cross-lingual word vector space. 
\end{itemize}

\paragraph{Implementation detail.} 
We first used the fastText model to embed the 36K English-German word pairs into vectors. 
Then we normalized each vector to unit $\ell_2$ norm and removed the per-dimension mean and standard deviation of the training pairs.

To build an end-to-end model, we followed the same intuition as DCCAE 
but instead of using the latent representation from the encoder to reconstruct the inputs, 
we used the outputs of proximal layer, which is a proximal approximation of latent representation from the encoder, to do the reconstruction. 
That is, the input reconstruction error was used as the ultimate objective.

We implemented the encoder (feature mapping $f$ and $g$) by using multilayer perceptrons with ReLU activation and the decoder by using a symmetric architecture of encoder. 
We tuned the hidden dimension $h$ for $f$ and $g$ among $\{0.1, 0.3, 0.5, 0.7, 0.9\} \times 300$,
the regularization parameter $\lambda$ from $\{0.001, 0.01, 0.1, 1, 10\}$,
and the depth and layer width from 1 to 4 and $\{256, 512, 1024, 2048\}$, respectively.
For optimization, we used SGD with momentum 0.99, 
a weight decay of 0.0005, 
and a learning rate 0.1 which was divided by 10 after 100 and 200 epochs. 

At test time, for numerical stability, we combined the word vectors from bilingual dictionary and the test set to build paired vocabulary for each language. 
We applied the same data preprocessing (normalize to unit norm, remove the mean/standard deviation of the training set) on test vocabularies (English and German word vectors).
Then we feed paired test vocabularies into the models and obtained the word representation of the test data. 
We projected the output of the proximal layer to the subspace where each paired word representation was maximally correlated. 
The projection matrices were calculated from the 36K training set through the standard CCA method. 
We computed the cosine similarity between the final word vectors in each pair, 
ordered the pairs by similarity, 
and computed the Spearman’s correlation between the model’s ranking and human's ranking.

\subsection{Adversarial Training in Recurrent Neural Network}
\label{sec:app_prox_lstm}

Here we include more details on the experiment of adversarial training in recurrent neural network as described in Section \ref{sec:exp_adversarial}. 

\paragraph{Datasets.} 
To demonstrate the effectiveness of using proximal mapping, we tested on four different sequence datasets. 
The Janpanese Vowels dataset \citep[JV][]{ShiKudToy99} contains time series data where nine male speakers uttered Japanese Vowels successively, 
and the task is to classify speakers. 
The Human Activity Recognition dataset \citep[HAR][]{AngGhiOne13} is used to classify a person's activity (sitting, walking, etc.) based on a trace of their movement using sensors. 
The Arabic Digits dataset \citep[AD,][]{HamBed10} contains time series corresponding to spoken Arabic digits by native speakers, and the task is to classify digits. 
IMDB \citep{MaaDalPha11} is a standard movie review dataset for sentiment classification. Details of the datasets are summarized in Table \ref{tab:dataset_summary_appendix}. 
The - is because IMDB is a text dataset, 
for which a 256-dimensional word embedding is learned.

\begin{table}[t!]
	\centering
	\caption{Summary of datasets for adversarial LSTM training}
	\label{tab:dataset_summary_appendix}
	\begin{tabular}{lcccccc}
		\toprule
		Dataset  & Training  & Test    & Median length  & Attributes & Classes \\ \midrule
		JV       & 225       & 370     & 15      & 12         & 9       \\
		HAR      & 6,127     & 2,974   & 128     & 9          & 6       \\
		AD       & 5,500     & 2,200   & 39      & 13         & 10      \\ 
		IMDB     & 25,000    & 25,000  &239      & -          & 2       \\\bottomrule
	\end{tabular}
\end{table}

\paragraph{Preprocessing.}
Normalization was the only preprocessing applied to all datasets. 
For those datasets that contain variable-length sequences, 
zero-padding was used to make all sequences have the same length as the longest sequence in a mini-batch. 
To reduce the effect of padding, 
we first sorted all sequences by length (except the IMDB dataset), so that sequences with similar length were assigned to the same mini-batch.

\paragraph{Baseline models.} 
To show the impact of applying proximal mapping on LSTM, we compared our method with two baselines. 
For JV, HAR and AD datasets, the base model structure was composed of a CNN layer, followed by an LSTM layer and a fully-connected layer. 
The CNN layer was constructed with kernel size 3, 8, 3 and contained 32, 64, 64 filters for JV, HAR, AD respectively. 
For the LSTM layer, the number of hidden units used in these three datasets are 64, 128, 64, respectively. 
This architecture was denoted as LSTM in Table \ref{tab:lstm_result}. 
For IMDB, following \citet{MiyDaiGoo17}, 
the basic model consisted of a word embedding layer with dimension 256, 
a single-layer LSTM with 1024 hidden units, 
and a hidden dense layer of dimension 30. 

On top of this basic LSTM structure, we compared two different adversarial training methods. AdvLSTM is the adversarial training method in \cite{MiyDaiGoo17}, 
which we reimplemented in PyTorch, and perturbation was added to the input of each LSTM layer. 
ProxLSTM denotes our method described in Section \ref{sec:lstm}, 
where the LSTM cell in the basic structure was replaced by our ProxLSTM cell. 
LSTM and AdvLSTM here correspond to ``Baseline'' and ``Adversarial'' in \citet{MiyDaiGoo17} respectively.

\paragraph{Training.}
For the JV, HAR, AD datasets,
we first trained the baseline LSTM to convergence, 
and then applied AdvLSTM and ProxLSTM as fine tunning,
where ADAM was used with learning rate $10^{-3}$ and weight decay $10^{-4}$. 
For IMDB, we first trained LSTM and AdvLSTM by following the settings in \cite{MiyDaiGoo17}, 
with an ADAM optimizer of learning rate $5\cdot 10^{-4}$ and exponential decay $0.9998$. 
Then the result of AdvLSTM was used to initialize the weights of ProxLSTM.
All settings were evaluated 10 times to report the mean and standard deviation.

\begin{figure*}[htbp!]
\centering
  \includegraphics[width=120mm, height=44mm]{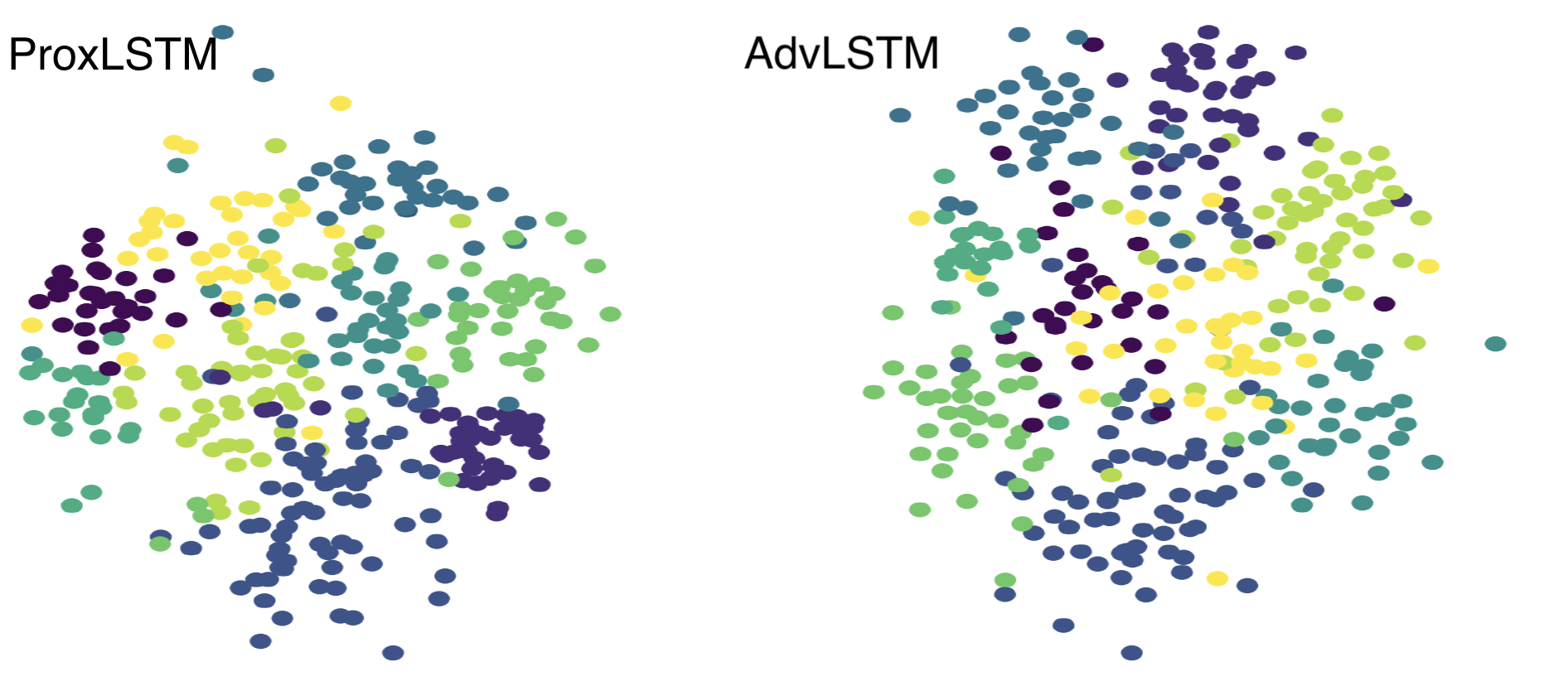}
  \caption{t-SNE embedding of the JV dataset}
  \label{fig:JV_tsne}
  \vspace{3em}
\includegraphics[width=120mm, height=44mm]{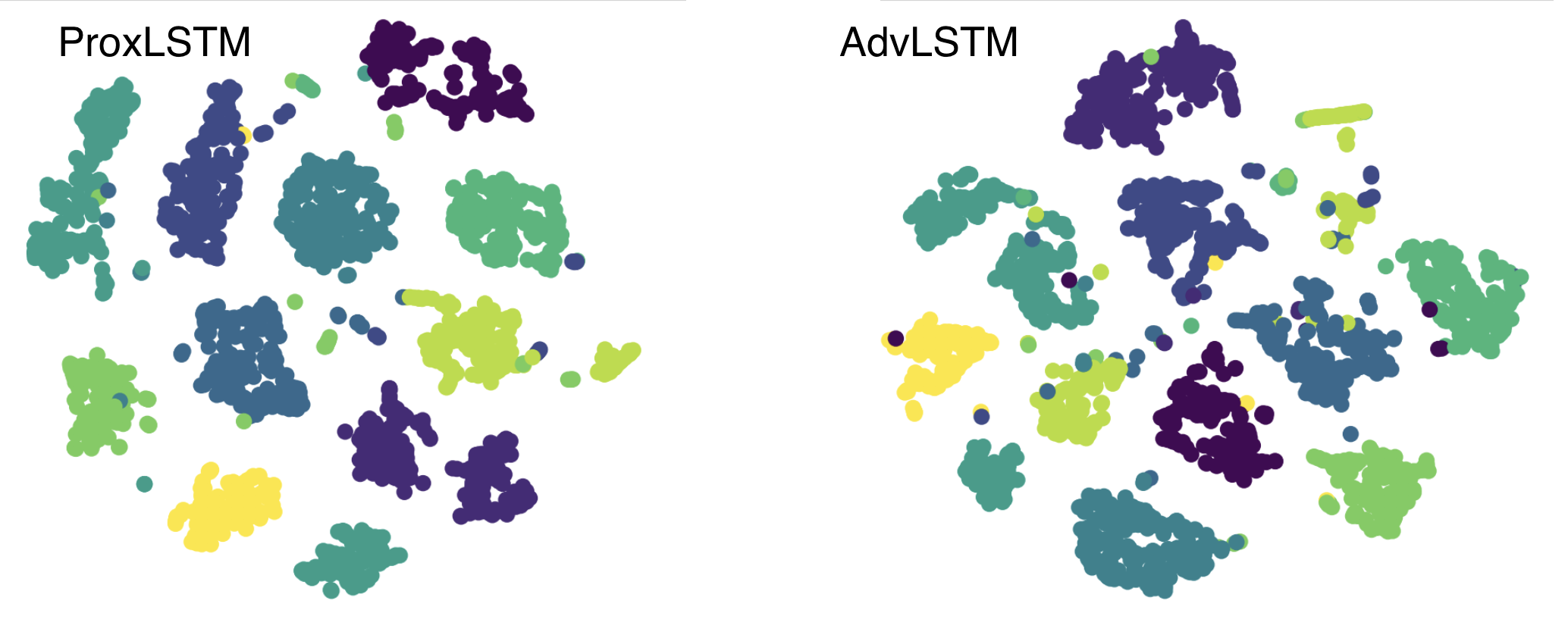}
  \caption{t-SNE embedding of the AD dataset}
  \label{fig:AD_tsne}
    \vspace{3em}
\includegraphics[width=120mm, height=44mm]{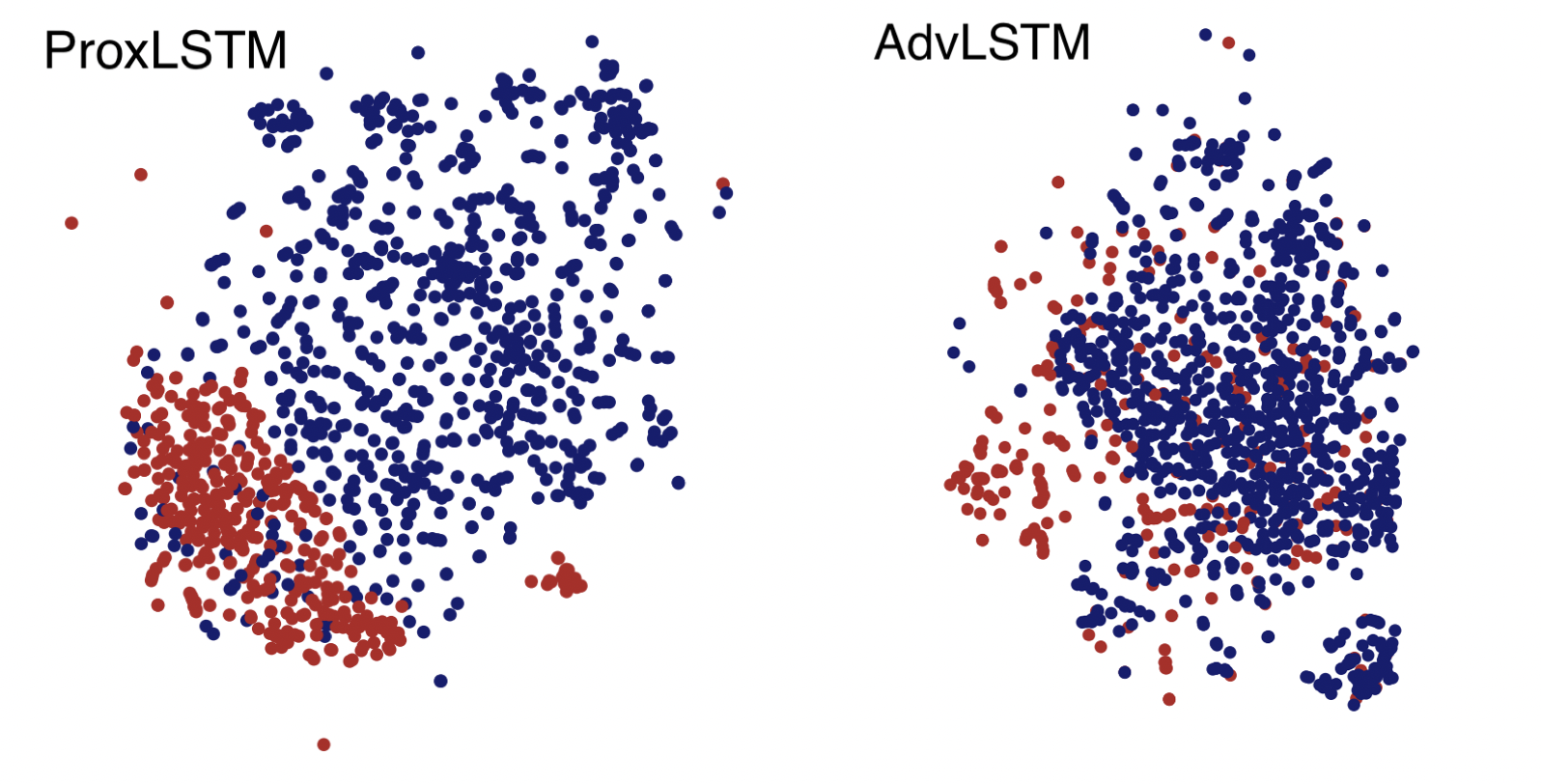}
    \caption{t-SNE embedding of the IMDB dataset}
    \label{fig:IMDB_tsne}

\end{figure*}

\paragraph{Results.} 
The test accuracies were summarized in Table \ref{tab:lstm_result}. 
Clearly, adversarial training improves the performance, 
and ProxLSTM even promotes the performance more than AdvLSTM. 
Figure \ref{fig:lstm_tsne} illustrates the t-SNE embedding of extracted features from the last time step's hidden state of HAR test set. 
Although ProxLSTM only improves upon AdvLSTM marginally in test accuracy, 
Figure \ref{fig:lstm_tsne} shows the embedded features from ProxLSTM cluster more compactly than those of AdvLSTM (e.g. the yellow class). 
The t-SNE plot of other datasets are available in Figures \ref{fig:JV_tsne}, \ref{fig:AD_tsne} and \ref{fig:IMDB_tsne}. This further indicates that ProxLSTM can learn better latent representation than AdvLSTM by applying proximal mapping.

\end{document}